\documentclass[11pt]{article}

\usepackage[final]{acl}
\usepackage{kotex}
\usepackage{times}
\usepackage{latexsym}
\usepackage[T1]{fontenc}
\usepackage[utf8]{inputenc}
\usepackage{microtype}
\usepackage{inconsolata}
\usepackage{graphicx}
\usepackage{titlesec} 
\usepackage{caption}  
\usepackage{amsmath}
\usepackage{dsfont}
\usepackage{subcaption}
\usepackage{tabularx}
\usepackage{multirow}

\title{Understanding Cross-Modal Inconsistency through Representation Subspaces in Omni Models}

\author{Seonghyeon Cho, Chanjun Park\textsuperscript{\textdagger}
\\
\\Soongsil University
\\
   \texttt{chosh040@soongsil.ac.kr, chanjun.park@ssu.ac.kr}
}

\begin{document}
\maketitle

\begingroup
\renewcommand{\thefootnote}{\textdagger}
\footnotetext{Corresponding authors.}
\endgroup

\begin{abstract}
Omni Models process audio, vision, and text within a unified language backbone, yet often produce different answers for semantically equivalent inputs across modalities. Existing representation-level analyses do not distinguish between information that predicts whether disagreement will occur and the representation whose perturbation affects the model's answer. In this paper, we estimate a low-dimensional Answer Subspace from answer-position representations and use it to separate answer-related components from the remaining representation. Across three Omni Models and two multimodal reasoning tasks, information predictive of cross-modal inconsistency lies mainly outside the Answer Subspace, concentrating in a reproducible two-dimensional representation that is not explained by a fixed modality offset. Matched interventions reveal distinct behavioral roles: perturbing the Answer Subspace substantially changes the model's predictions, whereas perturbing the disagreement representation, like a random subspace of the same dimensionality, leaves them almost unchanged. Our key finding is that cross-modal disagreement can be induced by misaligning modality-specific answer representations: perturbing the Answer Subspace for only one modality induces disagreement, whereas applying the same perturbation to both largely preserves agreement. Thus, the representation that predicts disagreement is distinct from the answer-level representation through which disagreement can be induced, showing that linear decodability does not imply behavioral efficacy.
\end{abstract}

\section{Introduction}\label{int}
Recent Multimodal Large Language Models (MLLMs) have evolved beyond image–text understanding toward Omni Models that jointly process diverse modalities, including audio, and support any-to-any multimodal interaction. Omni Models integrate information extracted by modality-specific encoders within a shared language backbone to generate responses to heterogeneous sensory inputs. This capability has positioned Omni Models as promising foundation models for applications such as embodied agents and physical AI, which must understand and reason about the real world from multiple sensory signals. Such applications, in particular, require models to make consistent decisions even when the same environment or state is observed through different modalities. In practice, however, Omni Models often exhibit cross-modal inconsistency, producing different answers when semantically equivalent information is presented through different modalities, thereby undermining model reliability.

To address this issue, prior work has focused on constructing benchmarks for quantitatively evaluating cross-modal consistency and on mitigating inconsistency through training strategies such as modality alignment and preference optimization. These studies have established an important foundation by systematically measuring modality-dependent behavioral differences and demonstrating that consistency can be improved through additional training. However, because these approaches primarily evaluate models in terms of final-answer accuracy, agreement, or performance gains after training, they provide limited insight into how inputs from different modalities are transformed and computed within the language backbone to produce divergent answers.

Recent studies have therefore begun to diagnose cross-modal inconsistency by examining internal model states, including hidden representations, attention patterns, and activations. Nevertheless, representations derived from different modalities jointly encode modality identity, semantic content, and answer-related information. Consequently, directly comparing such representations makes it difficult to isolate the components involved in answer formation.

In this work, we analyze cross-modal inconsistency at the representation level by focusing on answer-position representations in the shared language backbone of Omni Models. We first estimate a low-dimensional subspace from which the model's current prediction can be linearly decoded and use it as a reference coordinate system. Based on this representation, we investigate: (i) where information predictive of modality-dependent disagreement resides and how low-dimensional it is; (ii) what this representation reflects, in particular whether it corresponds to a fixed modality difference or an item-specific component; and (iii) how it differs from the answer representation under matched interventions.

Our results show that information predictive of cross-modal disagreement lies mainly outside the Answer Subspace, forms a reproducible two-dimensional representation, and is not explained by a fixed modality offset. Under matched interventions, however, the two subspaces behave differently: perturbing the Answer Subspace changes the model's prediction, whereas perturbing the disagreement subspace leaves it essentially unchanged. Moreover, perturbing the Answer Subspace for only one modality reliably induces cross-modal disagreement, while the same perturbation applied to both modalities does not.

Our main contributions are as follows:

\begin{itemize}
    \item We estimate a low-dimensional subspace at the answer position and use it to separate representation components carrying the model's answer from the remaining representation.

    \item Using this separation, we show that information predictive of cross-modal inconsistency lies outside the Answer Subspace and is concentrated in a reproducible two-dimensional representation that is not explained by a fixed modality offset.

    \item We show that, under matched additive interventions, the disagreement-predictive representation has no greater behavioral effect than random directions, whereas asymmetric perturbations of the Answer Subspace are sufficient to induce cross-modal disagreement.
\end{itemize}

\section{Related Work}\label{rw}
\subsection{Cross-modal Consistency in Multimodal Language Models}
Multimodal language models can produce different answers when semantically equivalent information is presented through different modalities. Such \emph{cross-modal inconsistency} has been observed in vision–language models~\cite{zhang2025crossmodalinformationflowmultimodal,tang2025seamsemanticallyequivalentmodalities}, across text, image, and mixed-modal inputs~\cite{vansprang2026contentdifferentanswerscrossmodal}, in omni-modal settings~\cite{chen2024omnixrevaluatingomnimodalitylanguage}, and in complex multimodal documents~\cite{yan2025multimodalinconsistencyreasoningmmir}. Related benchmarks diagnose visual conflicts and hallucinations~\cite{pezeshkpour2025mixedsignalsdecodingvlms,liu2025insightsightexploringvisionknowledge,li2023evaluating,guan2024hallusionbenchadvanceddiagnosticsuite}, and decoding and training strategies have been proposed to reduce such inconsistencies~\cite{leng2024mitigating,zhang2025debiasingmultimodallargelanguage}.

At the representation level, linear probing, representational similarity analysis, and centered kernel alignment have been used to compare what is encoded across modalities~\cite{alain2016understanding,kriegeskorte2008representational,kornblith2019similarityneuralnetworkrepresentations}, alongside work on shared embedding spaces~\cite{radford2021learningtransferablevisualmodels,girdhar2023imagebindembeddingspacebind,zhu2024languagebindextendingvideolanguagepretraining}. These approaches compare representations as a whole, and therefore do not distinguish the components that carry the model's answer from those that do not.

\subsection{Mechanistic Interpretability and Decision Representations}
Mechanistic interpretability relates model behavior to internal representations through targeted interventions. Causal tracing and activation patching measure the influence of specific activations by replacing or restoring them~\cite{meng2022locating,wang2022interpretabilitywildcircuitindirect,conmy2023towards,zhang2023towards}, while iterative nullspace projection and amnesic probing remove linearly decodable information and examine the resulting behavioral change~\cite{ravfogel2020inlp,elazar2021amnesicprobingbehavioralexplanation}. Decoding studies ask what decision-related information the hidden states already carry: Tuned Lens and Future Lens estimate future output distributions from intermediate states~\cite{belrose2023eliciting,pal2023future}, and probing studies examine whether factual knowledge, truthfulness, and confidence are linearly decodable~\cite{burns2024discoveringlatentknowledgelanguage,azaria2023internal,marks2023geometry,burger2024truth,kadavath2022language}.

Prior work has not established the internal mechanism by which semantically equivalent inputs lead to different answers across modalities. We address this gap by locating disagreement-predictive representations relative to answer-related ones and testing, through intervention, their roles in producing cross-modal inconsistency.

\section{Method}\label{mtd}
\subsection{Preliminaries}\label{mtd:pre}
\paragraph{Multimodal Large Language Models.}
MLLMs extend autoregressive language models to process multiple modalities, including images and audio, in addition to text. Inputs from each modality are transformed by modality-specific encoders into representations that can be processed by the language backbone, after which inference is performed within the shared language backbone~\cite{alayrac2022flamingo, liu2023visual, chu2024qwen2}.

Let $\mathcal{M}$ denote the set of modalities, and let $m\in\mathcal{M}$ denote a specific input modality. We denote the input representation of modality $m$ to the language backbone by $\mathbf{I}^{(m)}$, and the hidden representation of the $t$-th token at the $l$-th layer by $\mathbf{h}^{(m,l)}_t$.

\paragraph{Omni Models.}
An Omni Model is an MLLM extended to process diverse input and output modalities, including images, audio, and text, within a single model~\cite{zhan2024anygpt, xu2025qwen3}. Although the encoder architecture and input integration mechanism vary across models, this work focuses on Omni Models in which inputs from different modalities are jointly processed by a shared language backbone.

Each modality may undergo a distinct encoding and preprocessing pipeline, but the hidden representations produced by the language backbone lie in a common representation space. We therefore directly compare hidden representations from different input modalities at both the layer and token levels within this shared space.

\subsection{Problem Formulation}\label{mtd:pf}
We present each item to an Omni Model through semantically equivalent inputs in two different modalities and require the model to generate a single answer choice. Rather than focusing on the final output itself, we analyze how modality-dependent behavioral differences are reflected in the answer-position representation at the point where the answer is generated.

At layer $l$, we define the hidden representation at the answer-generation position as

\begin{equation}\label{eq:hans}
    \mathbf h_{\mathrm{ans}}^{(m,l)}
    =
    \mathbf h_{t_{\mathrm{ans}}}^{(m,l)}
    \in
    \mathds{R}^{d},
\end{equation}

where $t_{\mathrm{ans}}$ denotes the position at which the logits for the answer choice are computed, $l$ indexes the layer of the language backbone, $m$ denotes the input modality, and $d$ is the hidden dimension. Because the number of layers differs across models, we report layer depth in normalized form.  The representation extraction procedure is described in Appendix~\ref{app:extract}.

To connect this representation to model behavior, we define the prediction for item $i$ as

\begin{equation}\label{eq:pred}
    \hat y_i^{(m)}
    =
    f\!\left(x_i^{(m)},\mathbf c_i\right),
    \qquad
    m\in\{\mathrm a,\mathrm v\}.
\end{equation}

Here, $x_i^{(m)}$ denotes item $i$ presented in modality $m$, and $\mathbf c_i=(c_i^{A},c_i^{B},c_i^{C},c_i^{D})$ denotes the four answer choices shared across the two modalities. The prediction $\hat y_i^{(m)}\in\{A,B,C,D\}$ is the answer-choice label generated by the model for the corresponding input.

We define cross-modal disagreement as the case in which the model produces different predictions across the two modalities:

\begin{equation}\label{eq:disagreement}
    z_i
    =
    \mathds 1
    \!\left[
        \hat y_i^{(\mathrm a)}
        \neq
        \hat y_i^{(\mathrm v)}
    \right].
\end{equation}

In Eq.~\ref{eq:disagreement}, $z_i=0$ denotes agreement and $z_i=1$ denotes disagreement.

\subsection{Subspace Estimation}\label{mtd:subspace}
\paragraph{Answer Subspace Probe.}
The answer-position representation contains not only information associated with the model's prediction but also variation arising from input format and modality-specific encoding. To isolate the components associated with the model's answer, we train a multinomial logistic regression probe at each layer using answer-position representations from both modalities. The prediction target is the answer-choice label $\hat y_i^{(m)}$ generated by the model for each input.

Let $W^{(l)}\in\mathds R^{d\times4}$ denote the coefficient matrix learned at layer $l$. Since softmax predictions are invariant to adding the same constant to all class logits, the effective rank of $W^{(l)}$ is at most three. We define the Answer Subspace as

\begin{equation}\label{eq:readout-subspace}
    \mathcal S_{\mathrm{ans}}^{(l)}
    =
    \mathrm{Col}\!\left(W^{(l)}\right).
\end{equation}

We obtain an orthonormal basis $Q_{\mathrm{ans}}^{(l)}\in\mathds R^{d\times3}$ for $\mathcal S_{\mathrm{ans}}^{(l)}$ by QR decomposition and define the component orthogonal to this subspace as

\begin{equation}\label{eq:residual}
    \mathbf r^{(m,l)}
    =
    \left(
        I_d
        -
        Q_{\mathrm{ans}}^{(l)}
        Q_{\mathrm{ans}}^{(l)\top}
    \right)
    \mathbf h_{\mathrm{ans}}^{(m,l)}.
\end{equation}

This decomposition allows disagreement-predictive information to be localized relative to the Answer Subspace. The probe-training procedure and computation of $Q_{\mathrm{ans}}^{(l)}$ are described in Appendix~\ref{app:readout}.

\paragraph{Disagreement Subspace Probe.}
Because disagreement is defined between two modality conditions, each probe sample corresponds to one item and combines the residual representations from both conditions:

\begin{equation}\label{eq:disfeat}
    \mathbf x_i
    =
    \left[
        \mathbf r_i^{(\mathrm a,l)};\;
        \mathbf r_i^{(\mathrm v,l)};\;
        \mathbf r_i^{(\mathrm v,l)}
        -
        \mathbf r_i^{(\mathrm a,l)}
    \right]
    \in
    \mathds R^{3d}.
\end{equation}

A binary probe is trained to predict $z_i$ in Eq.~\ref{eq:disagreement}. Restricting the input to the orthogonal complement ensures that the recovered directions exclude those already used to decode the answer. The learned coefficient vector is grouped according to the three feature blocks in Eq.~\ref{eq:disfeat}, yielding $W_z^{(l)}\in\mathds R^{3\times d}$. We define the disagreement subspace using the leading two right singular vectors of $W_z^{(l)}$,

\[
    B_{\mathrm{dis}}^{(l)}
    \in
    \mathds R^{d\times2}.
\]

Although the probe is trained on paired features in $\mathds R^{3d}$, the resulting basis lies in the original single-state space $\mathds R^d$ and can therefore be applied to individual hidden states. Full estimation details and a comparison with the two-block alternative are provided in Appendices~\ref{app:dis} and~\ref{app:blocks}.

\subsection{Subspace Intervention}\label{mtd:interv}
Linear decodability alone does not establish that a representation contributes to model behavior. We therefore intervene directly on each subspace and measure the resulting changes in the model's prediction.

Let $B\in\mathds R^{d\times k}$ denote an orthonormal basis for the target subspace. We apply the shift intervention directly to the answer-position representation:

\begin{equation}\label{eq:interv}
    \tilde{\mathbf h}_{\mathrm{ans}}^{(m,l)}
    =
    \mathbf h_{\mathrm{ans}}^{(m,l)}
    +
    \alpha B\boldsymbol\delta.
\end{equation}

where $\boldsymbol\delta\in\mathds R^k$ denotes a direction in the target-subspace coordinates and $\alpha$ controls the intervention magnitude. All comparisons use the same layer, normalization, magnitude, and operation, differing only in the target subspace.

To test whether cross-modal disagreement can be induced, we additionally apply Eq.~\ref{eq:interv} asymmetrically to items on which the two modalities initially agree. We perturb the Answer Subspace for one modality while leaving the other unchanged, and compare this condition with applying the same perturbation to both modalities.

The construction and normalization of $\boldsymbol\delta$, random-subspace controls, the removal intervention, and additional implementation details are provided in Appendix~\ref{app:interv}.

\section{Analysis}\label{sec:results}

\subsection{Experimental Setup}
\paragraph{Models.}
We evaluate three Omni Models with shared language backbones: Qwen2.5-Omni~\cite{xu2025qwen3}, Gemma-4-E4B~\cite{team2026gemma}, and MiniCPM-o 4.5~\cite{cui2026minicpm}.

\paragraph{Data.}
We use the \emph{spatial} and \emph{external} task families from XModBench~\cite{wang2026xmodbench}. Each item is presented under matched audio and image conditions with identical questions, answer choices, and A--D assignments; only modality-specific prompt wording differs. The same items are evaluated for all three models, yielding six model--task conditions.

\paragraph{Evaluation.}
Probes are evaluated with item-grouped cross-validation, keeping the two modality conditions of each item in the same fold, and balanced accuracy is reported throughout. Condition sizes and intervention layers are provided in Appendix~\ref{app:percond}.

\begin{figure*}[t]
    \centering

    \begin{subfigure}[t]{0.45\textwidth}
        \centering
        \includegraphics[width=\linewidth]{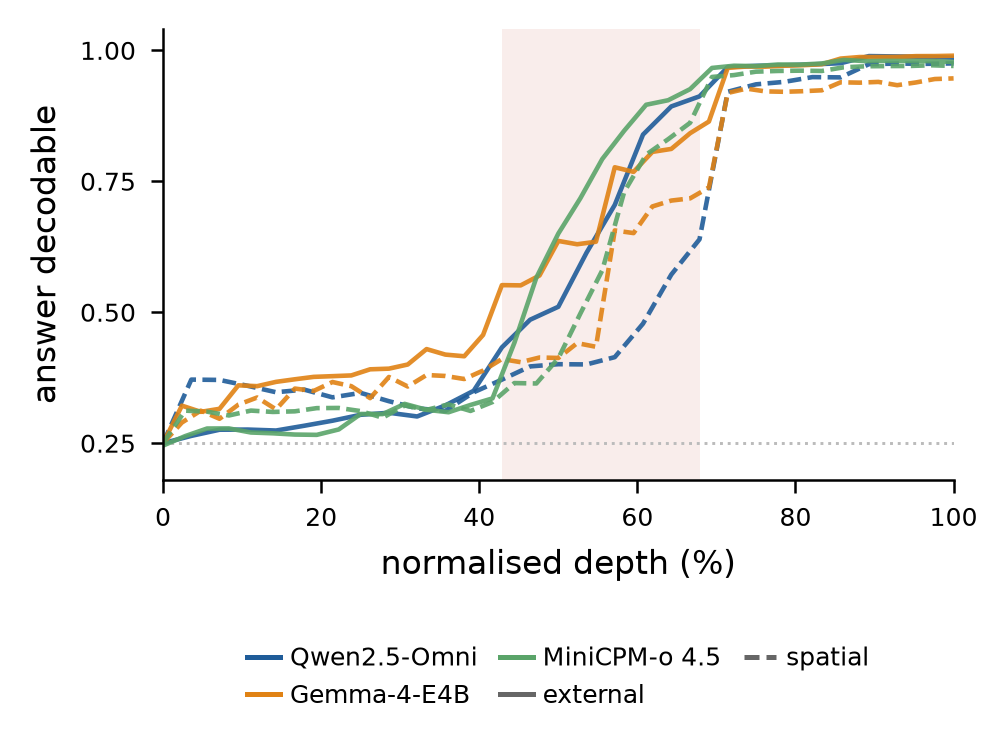}
        \caption{}
        \label{fig:fig1a}
    \end{subfigure}
    \hfill
    \begin{subfigure}[t]{0.45\textwidth}
        \centering
        \includegraphics[width=\linewidth]{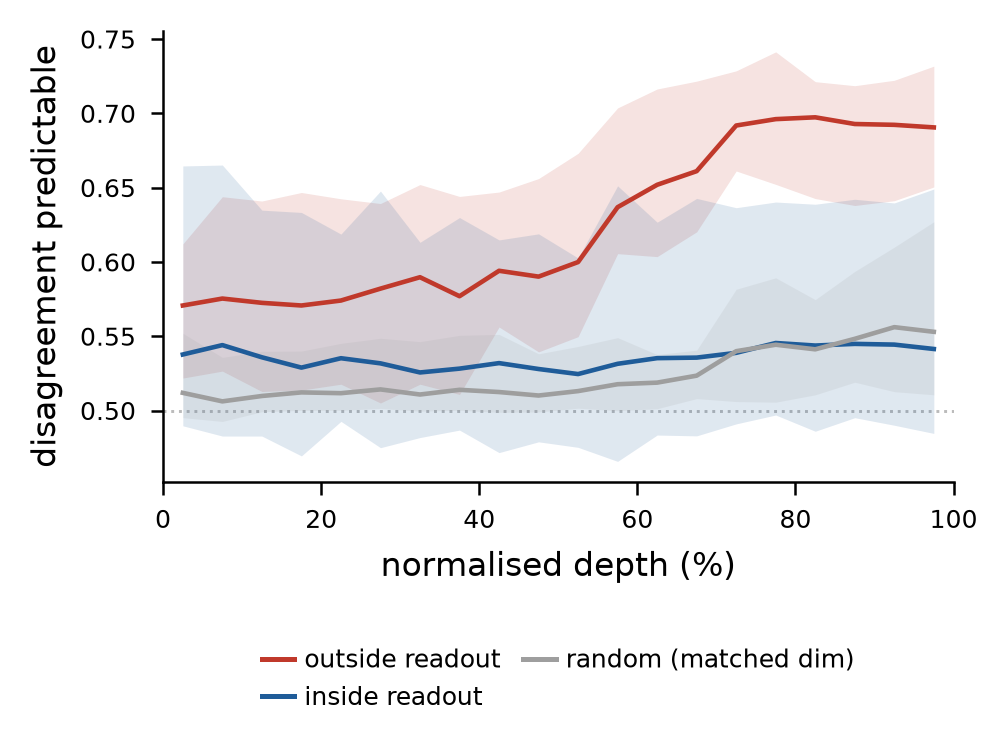}
        \caption{}
        \label{fig:fig1b}
    \end{subfigure}

    \caption{
        Layer-wise probe performance over normalized backbone depth. (a) Answer-probe accuracy shows that the model’s prediction becomes linearly decodable from the answer-position representation in the middle layers. (b) Disagreement-probe accuracy shows that disagreement-predictive information lies primarily outside the answer-related representation, with its orthogonal complement outperforming both the answer-related representation and matched random subspaces.
    }
    \label{fig:fig1}
\end{figure*}

\subsection{Where Is Disagreement-Predictive Information Encoded?}
\paragraph{Answer-related representation.}
Figure~\ref{fig:fig1a} shows the layer-wise accuracy of the answer probe introduced in Section~\ref{mtd:subspace}. Probe accuracy increases with depth and approaches perfect recovery in the final layer, first reaching the midpoint between chance and its final layer value around the middle of the backbone in every condition. Answer-related information therefore becomes reliably decodable well before the output, and the learned three-dimensional subspace provides a compact representation from which the model's predicted answer can be linearly decoded.

\paragraph{Disagreement-related representation.}
Figure~\ref{fig:fig1b} compares disagreement-probe performance using the answer-related representation, its orthogonal complement, and matched random subspaces. The orthogonal complement consistently outperforms both the answer-related representation and the random controls across all six conditions, with gains of $+0.042$--$0.225$ over the answer-related representation. These results indicate that information predictive of cross-modal disagreement resides primarily outside the answer-related representation, with the full per-condition breakdown provided in Appendix~\ref{app:loc}.

\begin{figure}[t]
    \centering
    \includegraphics[width=\linewidth]{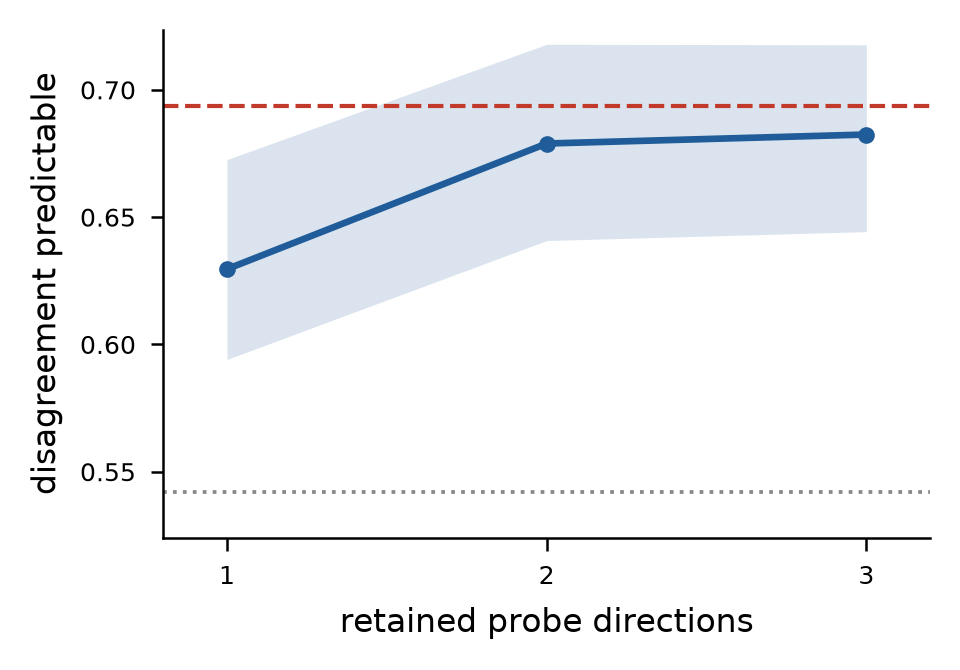}
    \caption{
        Disagreement-probe accuracy using the leading one, two, or three right singular vectors of the reshaped disagreement-probe coefficient matrix, selected within each training fold. Performance increases substantially from one to two directions, with little additional improvement from the third. Dashed and dotted lines denote the full hidden-state and Answer Subspace probes, respectively. Appendix~\ref{app:stab} reports per-condition results and bootstrap stability.
    }
    \label{fig:fig4}
\end{figure}

\paragraph{Low-dimensional approximation.}
To determine how many directions are needed to capture the dominant disagreement-predictive variation, we progressively retain the right singular vectors of the reshaped disagreement-probe coefficient matrix. Because this matrix has rank at most three, we evaluate one, two, and three retained directions. Figure~\ref{fig:fig4} shows that disagreement-probe performance increases substantially from one to two directions, with little or no consistent improvement from the third. We therefore retain the leading two directions as a compact representation for the subsequent analyses and interventions.

\subsection{What Do the Two Subspaces Represent?}
The preceding analyses identify two low-dimensional representation spaces: a three-dimensional space that decodes the model's prediction and a two-dimensional space that predicts cross-modal disagreement. Decodability alone, however, does not establish what these spaces represent. We therefore test whether the former tracks the model's current prediction and whether the latter reflects disagreement-specific structure rather than modality or answer identity.

\begin{figure}
    \centering
    \includegraphics[width=\linewidth]{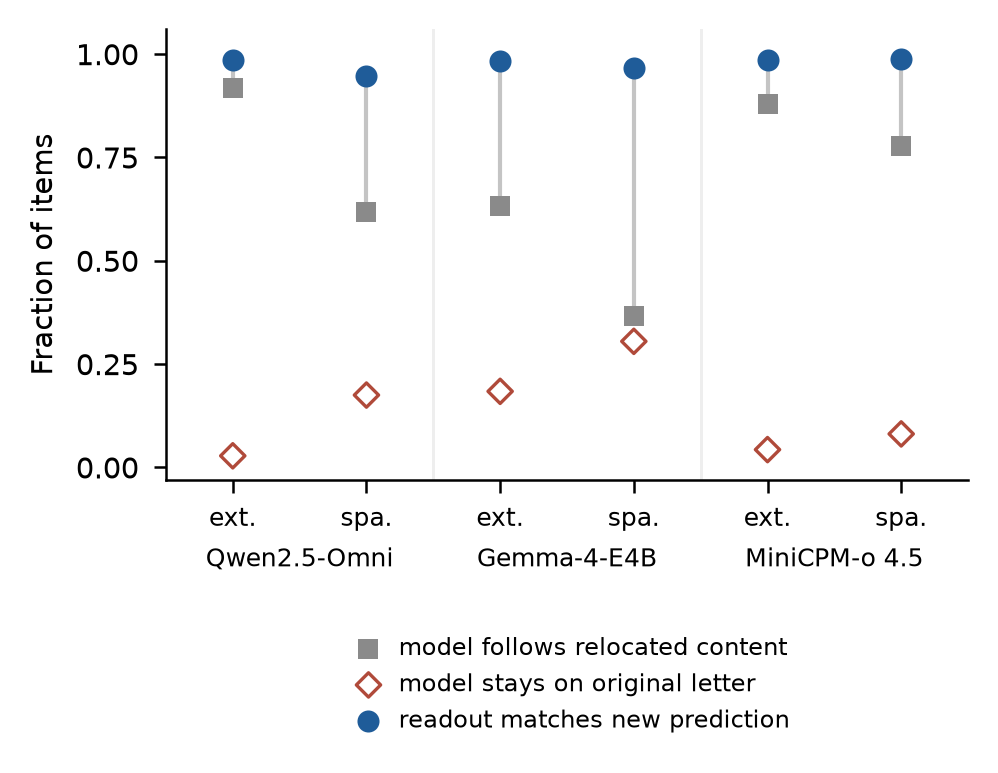}
    \caption{
        Answer decoding under option permutation. Despite substantial variation in model behavior, the decoded answer consistently matches the model's updated prediction. Squares, diamonds, and circles denote content following, letter retention, and decoded-answer agreement, respectively.
    }
    \label{fig:perm-control}
\end{figure}

\paragraph{Answer Subspace.}
The final-layer representation projected onto the answer-related space recovers the model's prediction with an accuracy of $0.937$--$0.997$, including items for which the two modalities produce different answers. Appendix~\ref{app:recover} reports the full per-condition results. Thus, the recovered coordinates preserve the prediction associated with each condition rather than collapsing the two modalities into a shared answer. We further test whether these coordinates track changes in the model's prediction by permuting the answer contents assigned to labels A--D.

As shown in Figure~\ref{fig:perm-control}, model behavior varies substantially across conditions, whereas the decoded answer continues to match the model's new prediction at $0.948$--$0.989$. The decoded answer therefore reflects the model's current prediction rather than either the original label assignment or the relocated answer content. Appendix~\ref{app:permtab} provides the detailed results. Together, these results validate the Answer Subspace as a faithful linear representation of the model's current answer.

\begin{figure}[t]
    \centering
    \includegraphics[width=\linewidth]{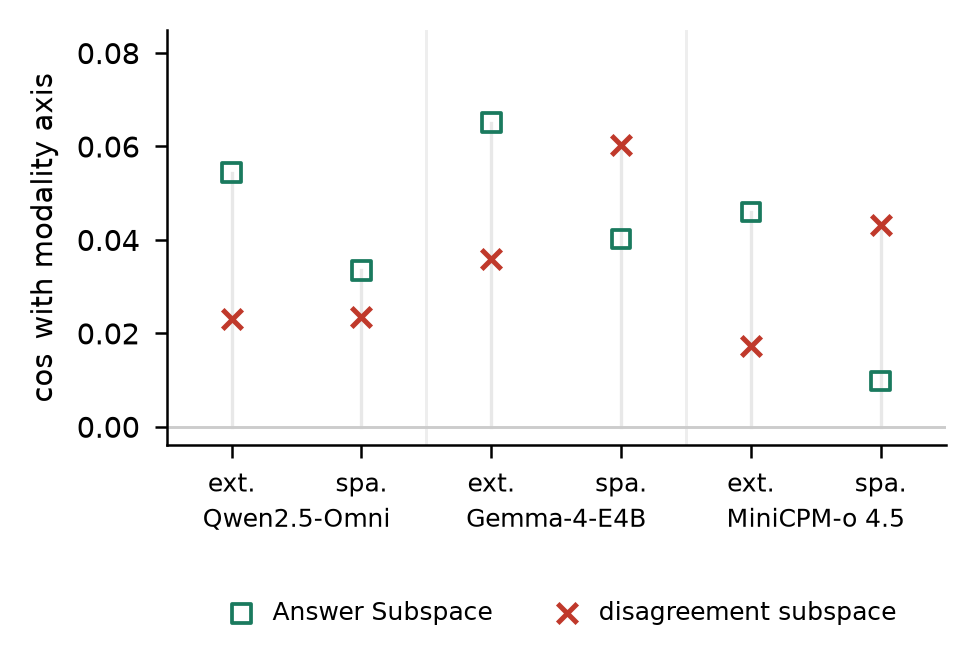}
    \caption{
        Alignment of the recovered subspaces with the modality-discriminating axis.
        Despite near-perfect modality separability, both subspaces are nearly
        orthogonal to this axis. Removing the mean modality offset has negligible
        effect on disagreement-probe accuracy. Appendix~\ref{app:geotab} reports
        the per-condition values.
    }
    \label{fig:modctrl}
\end{figure}

\begin{figure}[h]
    \centering
    \includegraphics[width=0.8\linewidth]{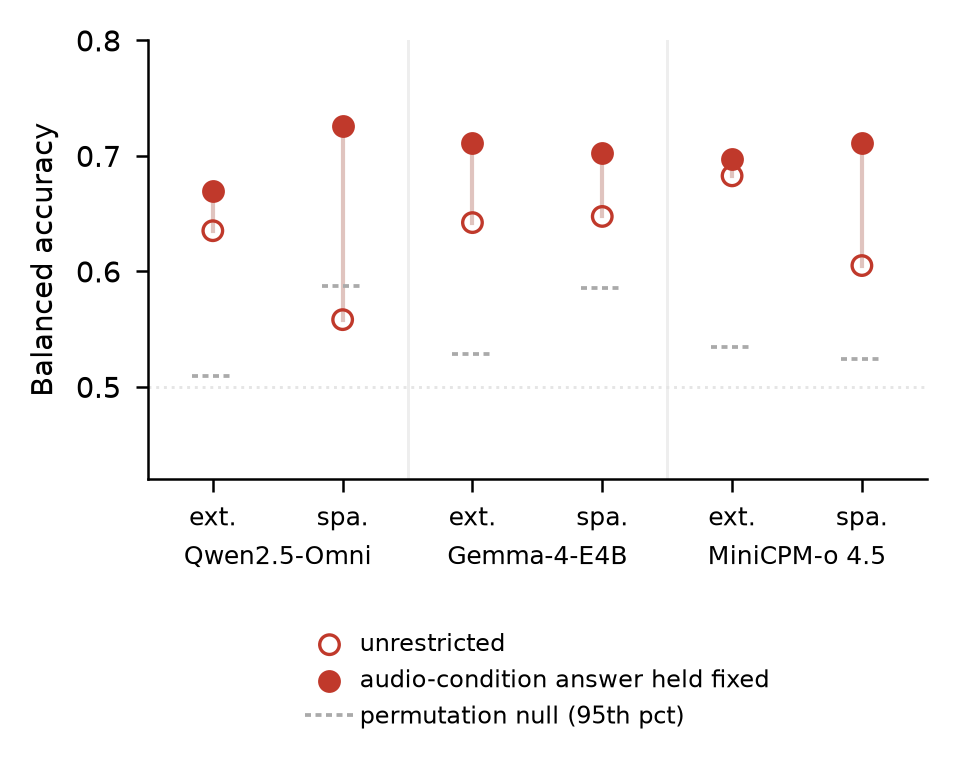}
    \caption{
        Disagreement-probe accuracy after controlling for the audio-condition
        prediction. Filled and open circles denote answer-controlled and
        unrestricted accuracy, respectively, and dashed ticks mark the
        label-permutation null. Appendix~\ref{app:ansmatch} reports the strata and
        sample sizes.
    }
    \label{fig:ansctrl}
\end{figure}

\paragraph{Disagreement-Predictive Subspace.}
Because the disagreement probe jointly reads both modality conditions, the recovered representation could reflect either a fixed modality difference or the answers produced under the two conditions. Figure~\ref{fig:modctrl} argues against the first explanation: although modality is almost perfectly decodable from the answer-position representation, the recovered directions are nearly orthogonal to the modality axis, and removing the mean modality offset has negligible effect on disagreement-probe accuracy.

Figure~\ref{fig:ansctrl} addresses the second by restricting the analysis to items with the same audio-condition prediction. Disagreement remains decodable above the label-permutation baseline in every condition and performs at least as well as in the unrestricted analysis. Together, these controls validate the recovered representation as a Disagreement-Predictive Subspace whose signal is not explained by modality or answer identity.

\begin{figure*}[t]
    \centering

    \begin{subfigure}[t]{0.32\textwidth}
        \centering
                \includegraphics[width=\linewidth]{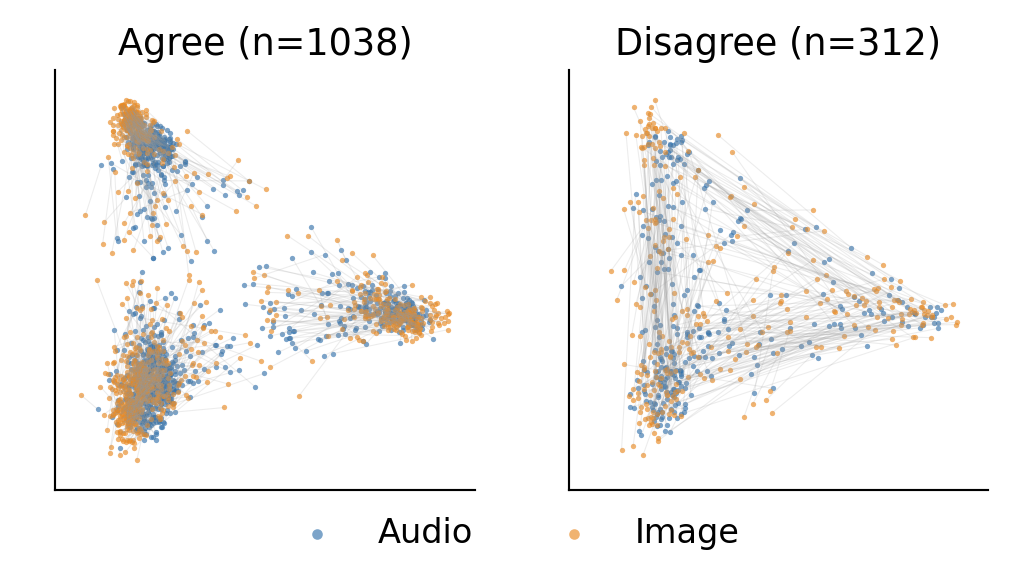}

        \caption{Qwen2.5-Omni (External)}
    \end{subfigure}
    \hfill
    \begin{subfigure}[t]{0.32\textwidth}
        \centering
                \includegraphics[width=\linewidth]{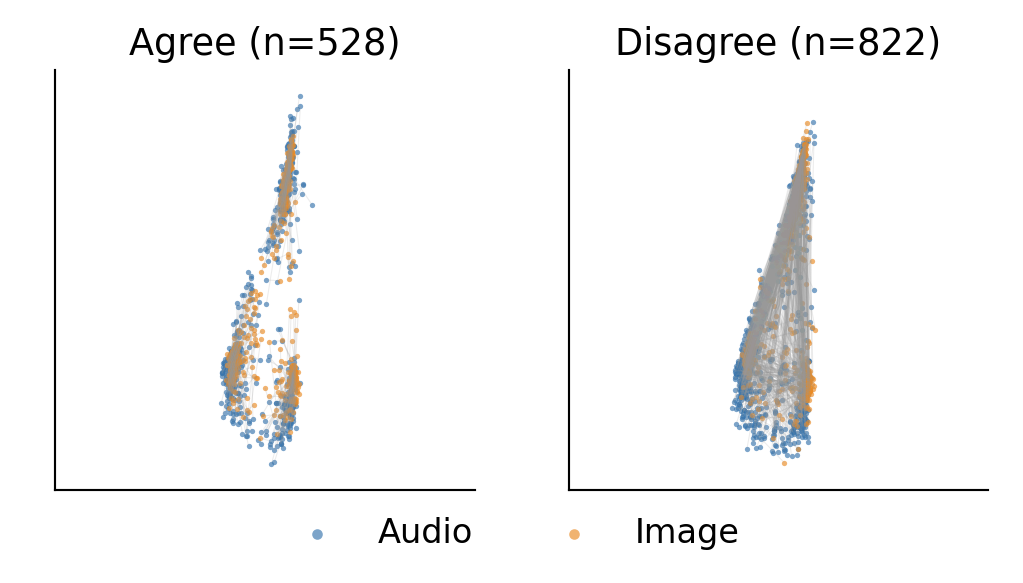}

        \caption{Gemma-4-E4B (External)}
    \end{subfigure}
    \hfill
    \begin{subfigure}[t]{0.32\textwidth}
        \centering
                \includegraphics[width=\linewidth]{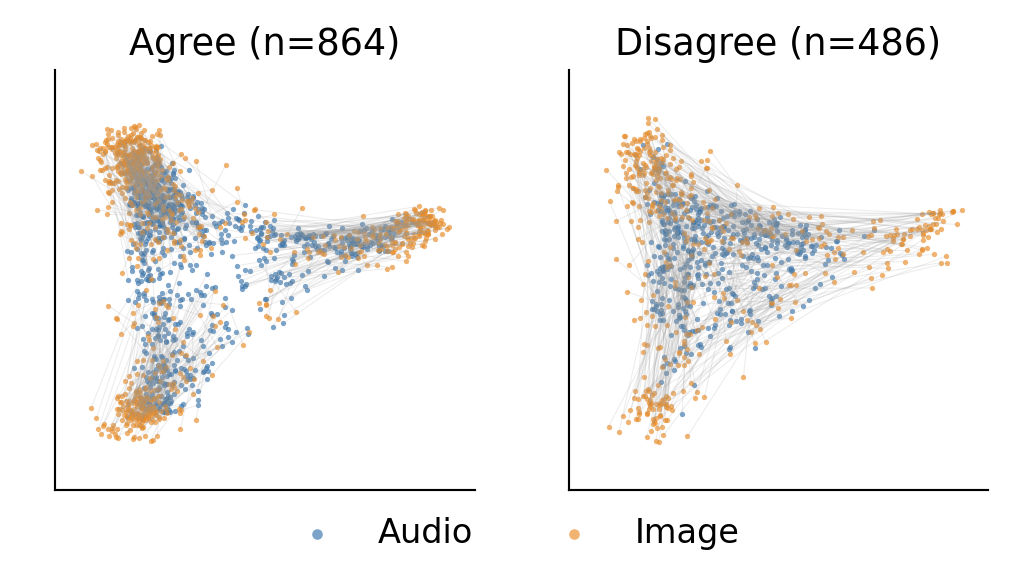}

        \caption{MiniCPM-o 4.5 (External)}
    \end{subfigure}

    \vspace{0.8em}

    \begin{subfigure}[t]{0.32\textwidth}
        \centering
                \includegraphics[width=\linewidth]{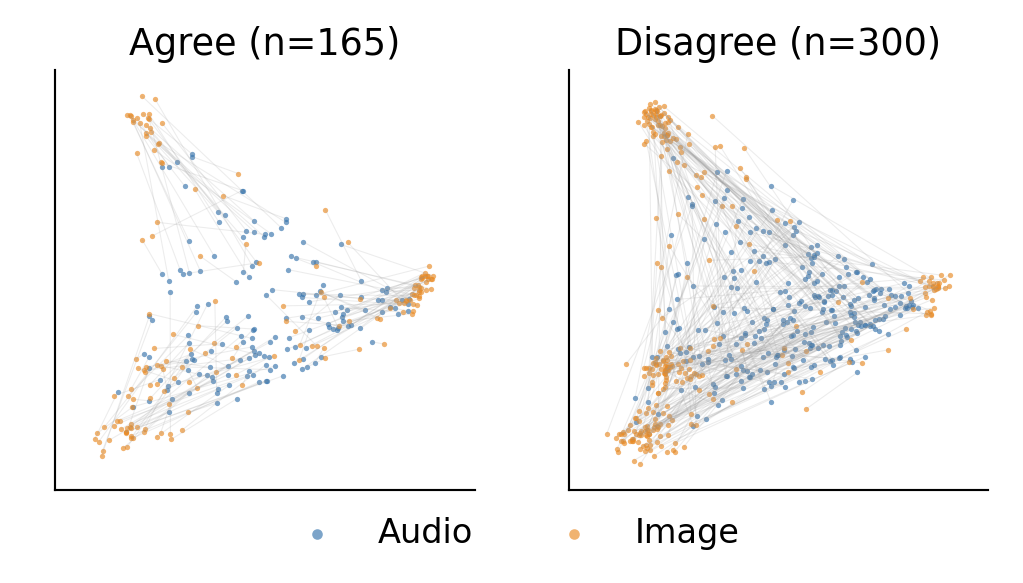}

        \caption{Qwen2.5-Omni (Spatial)}
    \end{subfigure}
    \hfill
    \begin{subfigure}[t]{0.32\textwidth}
        \centering
                \includegraphics[width=\linewidth]{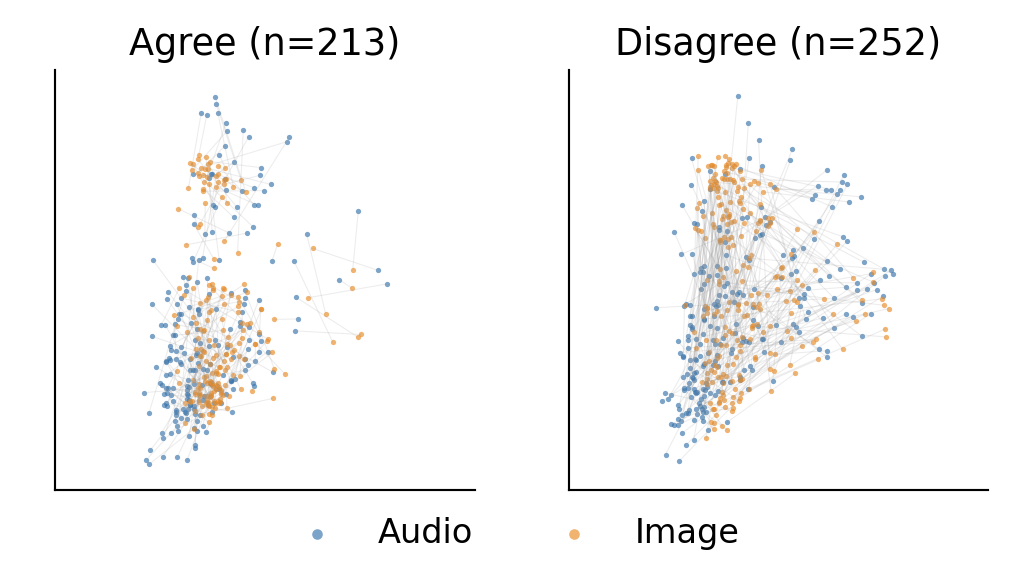}

        \caption{Gemma-4-E4B (Spatial)}
    \end{subfigure}
    \hfill
    \begin{subfigure}[t]{0.32\textwidth}
        \centering
                \includegraphics[width=\linewidth]{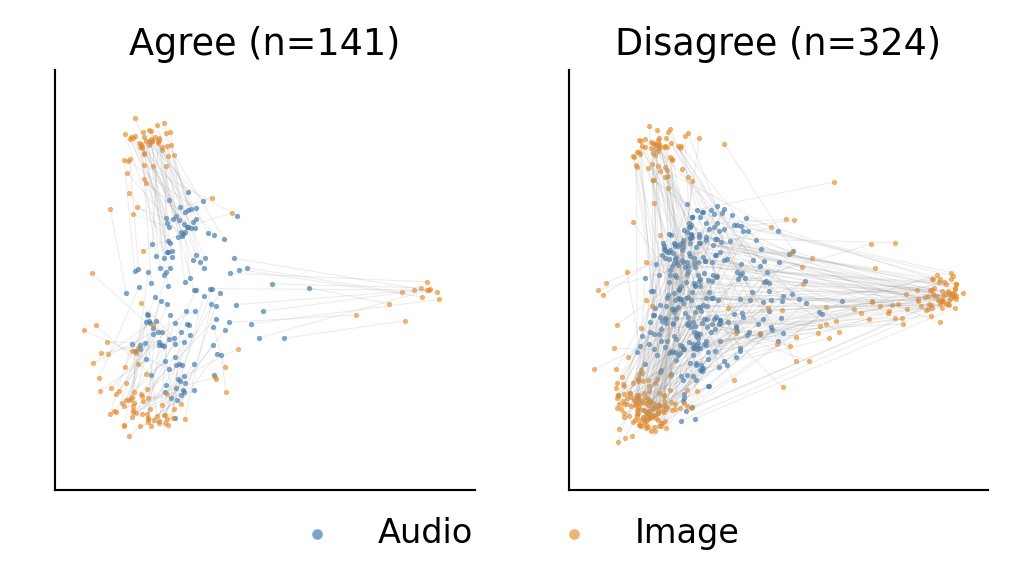}

        \caption{MiniCPM-o 4.5 (Spatial)}
    \end{subfigure}

    \caption{
        Paired audio and image representations in the Answer Subspace across six model--task conditions. Gray lines connect the two modality conditions of each item. Agreement pairs remain within the same answer region, whereas disagreement pairs separate across answer regions.
    }
    \label{fig:fig7}
\end{figure*}

\subsection{What Are the Behavioral Roles of the Two Subspaces?}
The preceding analyses identify two distinct representations: the Answer Subspace carries the answer produced under each modality condition, whereas the Disagreement-Predictive Subspace contains information associated with whether the two conditions will diverge. We first examine how naturally occurring disagreement is reflected in the Answer Subspace and then test the behavioral roles of the two subspaces.

Figure~\ref{fig:fig7} plots the two modality conditions of each item in the Answer Subspace. For agreement items, the two states lie close together within the same answer region. For disagreement items, they occupy different answer regions, and their median distance is $1.9$--$5.3$ times larger across the six conditions. After accounting for dimensionality, this excess separation is concentrated within the Answer Subspace rather than its orthogonal complement, by a factor of $2.9$--$8.6$. Appendices~\ref{app:pairs} and~\ref{app:geotab} report the full three-dimensional pair distances and modality-relative geometry. Thus, information predicting which items will diverge lies outside the Answer Subspace, whereas naturally occurring divergence is reflected as a separation between the two modality-specific answer states.

\begin{figure*}[t]
    \centering
    \includegraphics[width=0.9\linewidth]{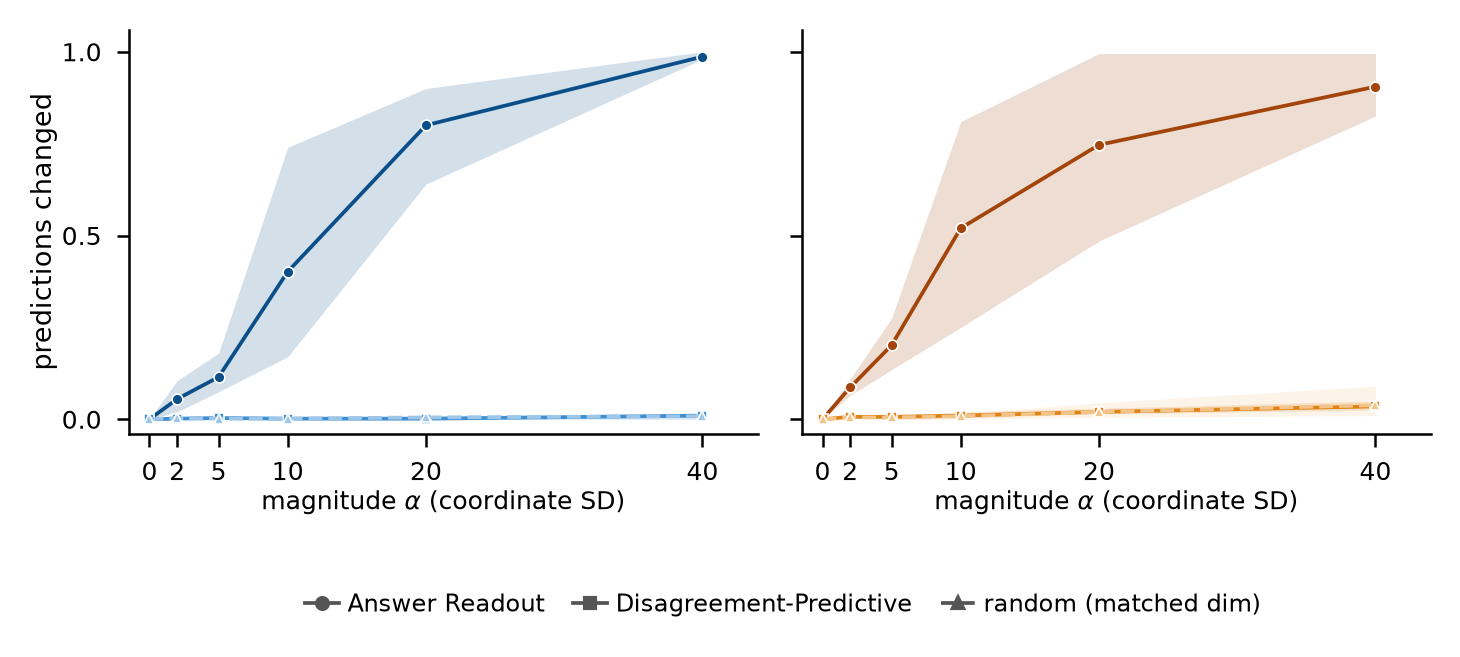}
    \caption{
        Prediction changes under matched interventions across shift magnitudes. Left and right panels show the \emph{external} and \emph{spatial} task families, respectively. Lines denote the mean across three models and shaded regions the range across models. Perturbations of the Answer Subspace produce a strong dose-dependent effect, whereas perturbations of the Disagreement-Predictive Subspace remain comparable to a random two-dimensional subspace drawn from the Answer Subspace complement.
    }
    \label{fig:fig8}
\end{figure*}

\paragraph{Behavioral effect.}
Figure~\ref{fig:fig8} compares matched interventions across a range of magnitudes. Perturbations within the Answer Subspace produce a clear dose-dependent behavioral effect, changing $0.83$--$1.00$ of predictions at the largest magnitude across the six conditions. By contrast, perturbations within the Disagreement-Predictive Subspace remain near baseline throughout the sweep and show no consistent advantage over a matched-dimensional random subspace. Appendix~\ref{app:dose} reports the full per-condition dose-response results.

The effect is also depth-dependent. Applying the same intervention before the answer becomes decodable reduces the effect of the Answer Subspace from $89$--$100$ to $0$--$3$ changed predictions out of $100$. Appendix~\ref{app:intervtab} reports the per-condition results. Behavioral sensitivity therefore emerges only after the Answer Subspace has formed, whereas the Disagreement-Predictive Subspace shows no comparable behavioral effect under matched interventions.

\begin{figure}[t]
    \centering
    \includegraphics[width=0.9\linewidth]{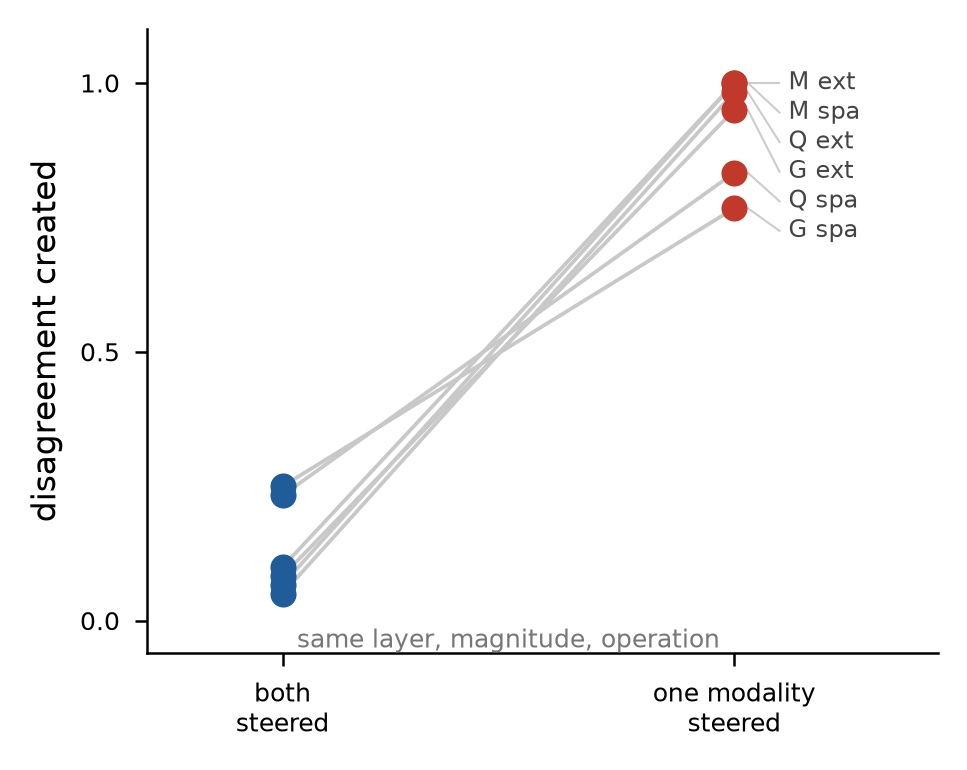}
    \caption{
        Disagreement induced by symmetric and asymmetric Answer Subspace interventions. Applying the perturbation to one modality produces substantially more disagreement than applying it to both under the same layer, magnitude, and operation.
    }
    \label{fig:fig10}
\end{figure}

\paragraph{Generating inconsistency.}
If manipulating the Disagreement-Predictive Subspace does not alter the model's prediction, what does? We address this question using items for which the two modalities initially produce the same answer. We perturb the Answer Subspace in only one modality while leaving the other unchanged, and compare this asymmetric intervention with applying the same perturbation to both modalities. As shown in Figure~\ref{fig:fig10}, perturbing only one modality induces substantially more disagreement than perturbing both. The asymmetric intervention creates disagreement in $0.767$--$1.000$ of the $60$ items per condition, whereas the symmetric intervention yields rates of only $0.050$--$0.250$, a gap of at least a factor of $3.1$ in every condition. No prediction changes are observed in the non-intervened modality. This comparison requires a magnitude at which the answer representation can be relocated; the asymmetric arm exceeds its control in all six conditions at the magnitude reported here and in five of six one step below, but not at smaller magnitudes (Appendix~\ref{app:asymsweep}).

\paragraph{Implications.}
These results distinguish the representation that predicts disagreement from the representation through which disagreement can be generated. The Disagreement-Predictive Subspace contains information associated with cross-modal divergence but has little behavioral effect under matched interventions. By contrast, the Answer Subspace does not predict whether disagreement will occur, yet perturbing it asymmetrically is sufficient to induce disagreement. Thus, linear decodability does not imply behavioral efficacy, and these two properties are distributed across distinct subspaces. Appendix~\ref{app:cohort} reports the cohort analysis, Appendix~\ref{app:removal} the removal interventions and their limitations, and Appendix~\ref{app:injdetail} verifies that the interventions reached their target states.

\section{Conclusion}
We investigated cross-modal inconsistency relative to the representation that encodes an Omni Model's answer. By estimating a low-dimensional Answer Subspace, we separated answer-related components from the remaining hidden state and localized disagreement-predictive information. This information lies outside the Answer Subspace and is concentrated in a reproducible two-dimensional subspace not explained by modality or answer identity. Matched interventions reveal distinct behavioral roles: perturbing the Disagreement-Predictive Subspace produces no greater effect than a matched random subspace, whereas perturbing the Answer Subspace changes predictions in a dose-dependent manner. Naturally occurring disagreement appears as separation between modality-specific states in the Answer Subspace, and asymmetric perturbation of this subspace is sufficient to induce disagreement. These results indicate that cross-modal inconsistency is expressed through misaligned modality-specific answer representations, while the representation predicting this mismatch does not directly control answer selection under the tested interventions. More broadly, a representation from which a behavioral property is decoded need not be the representation through which it is produced.

\section*{Limitations}
The disagreement-predictive coordinates are estimated from both modality conditions and therefore cannot flag disagreement before both predictions are available. Our use of \emph{predictive} refers to statistical covariation with the disagreement label rather than a deployable forecast. We do not establish what the recovered variation represents. Because disagreement items may differ systematically in difficulty or ambiguity, the recovered directions may partly reflect such factors. Holding the audio-condition prediction fixed removes one answer-related confound but not the remaining item-level variation. Our modality control establishes non-alignment rather than the absence of modality information. Modality remains partly decodable within the two-dimensional subspace, and removing its direction leaves too few dimensions to test the stronger claim. Finally, our experiments cover three Omni Models and two task families from a single benchmark. Although the models differ in depth, hidden dimension, and normalization, all place modality encoders before a decoder-only backbone. Generalization to other integration designs, modalities, task families, and training regimes remains untested. We also study only inference-time interventions and do not evaluate training-time mitigation.

\section*{Ethical considerations}
This work analyzes existing multimodal models and benchmark data and does not involve human participants, personal data, or sensitive content. The findings concern model behavior and should not be interpreted as guarantees of reliability in safety-critical applications. AI assistants were used to assist with code implementation, and all generated code was manually reviewed, validated, and approved by the authors before being used in the experiments.

\section*{Acknowledgement}
This work was supported by Korea Internet \& Security Agency(KISA) grant funded by the Korea government(PIPC) (No.RS-2026-25526342, Development of Technologies for Preventing Sensitive Information Inference and Risk Assessment in Foundation Model Operations). This work was also supported by the National Research Foundation of Korea(NRF) grant funded by the Korea government(MSIT) (RS-2026-25483747). In addition, this research was supported by Culture, Sports and Tourism R\&D Program through the Korea Creative Content Agency grant funded by the Ministry of Culture, Sports and Tourism in 2026 (Project Name: Develop AI agent technology to connect knowledge through public cultural facility-based discussion and communication, Project Number: RS-2026-25520645). Furthermore, this work was supported by Institute of Information \& communications Technology Planning \& Evaluation (IITP, AI Computing Support Project for R\&D) grant funded by the Korea government(MSIT) (High-Performance Research AI Computing Infrastructure Support at the 2 PFLOPS Scale, RS-2026-25505492). Lastly, we gratefully acknowledge VESSL for providing GPU resources.

\bibliography{custom}

@misc{zhang2025crossmodalinformationflowmultimodal,
      title={Cross-modal Information Flow in Multimodal Large Language Models}, 
      author={Zhi Zhang and Srishti Yadav and Fengze Han and Ekaterina Shutova},
      year={2025},
      eprint={2411.18620},
      archivePrefix={arXiv},
      primaryClass={cs.AI},
      url={https://arxiv.org/abs/2411.18620}, 
}

@misc{tang2025seamsemanticallyequivalentmodalities,
      title={SEAM: Semantically Equivalent Across Modalities Benchmark for Vision-Language Models}, 
      author={Zhenwei Tang and Difan Jiao and Blair Yang and Ashton Anderson},
      year={2025},
      eprint={2508.18179},
      archivePrefix={arXiv},
      primaryClass={cs.AI},
      url={https://arxiv.org/abs/2508.18179}, 
}

@misc{vansprang2026contentdifferentanswerscrossmodal,
      title={Same Content, Different Answers: Cross-Modal Inconsistency in MLLMs}, 
      author={Angela van Sprang and Laurens Samson and Ana Lucic and Erman Acar and Sennay Ghebreab and Yuki M. Asano},
      year={2026},
      eprint={2512.08923},
      archivePrefix={arXiv},
      primaryClass={cs.AI},
      url={https://arxiv.org/abs/2512.08923}, 
}

@misc{chen2024omnixrevaluatingomnimodalitylanguage,
      title={OmnixR: Evaluating Omni-modality Language Models on Reasoning across Modalities}, 
      author={Lichang Chen and Hexiang Hu and Mingda Zhang and Yiwen Chen and Zifeng Wang and Yandong Li and Pranav Shyam and Tianyi Zhou and Heng Huang and Ming-Hsuan Yang and Boqing Gong},
      year={2024},
      eprint={2410.12219},
      archivePrefix={arXiv},
      primaryClass={cs.AI},
      url={https://arxiv.org/abs/2410.12219}, 
}

@misc{yan2025multimodalinconsistencyreasoningmmir,
      title={Multimodal Inconsistency Reasoning (MMIR): A New Benchmark for Multimodal Reasoning Models}, 
      author={Qianqi Yan and Yue Fan and Hongquan Li and Shan Jiang and Yang Zhao and Xinze Guan and Ching-Chen Kuo and Xin Eric Wang},
      year={2025},
      eprint={2502.16033},
      archivePrefix={arXiv},
      primaryClass={cs.CL},
      url={https://arxiv.org/abs/2502.16033}, 
}

@misc{pezeshkpour2025mixedsignalsdecodingvlms,
      title={Mixed Signals: Decoding VLMs' Reasoning and Underlying Bias in Vision-Language Conflict}, 
      author={Pouya Pezeshkpour and Moin Aminnaseri and Estevam Hruschka},
      year={2025},
      eprint={2504.08974},
      archivePrefix={arXiv},
      primaryClass={cs.AI},
      url={https://arxiv.org/abs/2504.08974}, 
}

@misc{liu2025insightsightexploringvisionknowledge,
      title={Insight Over Sight: Exploring the Vision-Knowledge Conflicts in Multimodal LLMs}, 
      author={Xiaoyuan Liu and Wenxuan Wang and Youliang Yuan and Jen-tse Huang and Qiuzhi Liu and Pinjia He and Zhaopeng Tu},
      year={2025},
      eprint={2410.08145},
      archivePrefix={arXiv},
      primaryClass={cs.CL},
      url={https://arxiv.org/abs/2410.08145}, 
}

@inproceedings{li2023evaluating,
  title={Evaluating object hallucination in large vision-language models},
  author={Li, Yifan and Du, Yifan and Zhou, Kun and Wang, Jinpeng and Zhao, Xin and Wen, Ji-Rong},
  booktitle={Proceedings of the 2023 conference on empirical methods in natural language processing},
  pages={292--305},
  year={2023}
}

@misc{guan2024hallusionbenchadvanceddiagnosticsuite,
      title={HallusionBench: An Advanced Diagnostic Suite for Entangled Language Hallucination and Visual Illusion in Large Vision-Language Models}, 
      author={Tianrui Guan and Fuxiao Liu and Xiyang Wu and Ruiqi Xian and Zongxia Li and Xiaoyu Liu and Xijun Wang and Lichang Chen and Furong Huang and Yaser Yacoob and Dinesh Manocha and Tianyi Zhou},
      year={2024},
      eprint={2310.14566},
      archivePrefix={arXiv},
      primaryClass={cs.CV},
      url={https://arxiv.org/abs/2310.14566}, 
}

@inproceedings{leng2024mitigating,
  title={Mitigating object hallucinations in large vision-language models through visual contrastive decoding},
  author={Leng, Sicong and Zhang, Hang and Chen, Guanzheng and Li, Xin and Lu, Shijian and Miao, Chunyan and Bing, Lidong},
  booktitle={Proceedings of the IEEE/CVF Conference on Computer Vision and Pattern Recognition},
  pages={13872--13882},
  year={2024}
}

@misc{zhang2025debiasingmultimodallargelanguage,
      title={Debiasing Multimodal Large Language Models via Noise-Aware Preference Optimization}, 
      author={Zefeng Zhang and Hengzhu Tang and Jiawei Sheng and Zhenyu Zhang and Yiming Ren and Zhenyang Li and Dawei Yin and Duohe Ma and Tingwen Liu},
      year={2025},
      eprint={2503.17928},
      archivePrefix={arXiv},
      primaryClass={cs.CV},
      url={https://arxiv.org/abs/2503.17928}, 
}

@article{alain2016understanding,
  title={Understanding intermediate layers using linear classifier probes},
  author={Alain, Guillaume and Bengio, Yoshua},
  journal={arXiv preprint arXiv:1610.01644},
  year={2016}
}

@article{kriegeskorte2008representational,
  title={Representational similarity analysis-connecting the branches of systems neuroscience},
  author={Kriegeskorte, Nikolaus and Mur, Marieke and Bandettini, Peter A},
  journal={Frontiers in systems neuroscience},
  volume={2},
  pages={249},
  year={2008},
  publisher={Frontiers}
}

@misc{kornblith2019similarityneuralnetworkrepresentations,
      title={Similarity of Neural Network Representations Revisited}, 
      author={Simon Kornblith and Mohammad Norouzi and Honglak Lee and Geoffrey Hinton},
      year={2019},
      eprint={1905.00414},
      archivePrefix={arXiv},
      primaryClass={cs.LG},
      url={https://arxiv.org/abs/1905.00414}, 
}

@misc{radford2021learningtransferablevisualmodels,
      title={Learning Transferable Visual Models From Natural Language Supervision}, 
      author={Alec Radford and Jong Wook Kim and Chris Hallacy and Aditya Ramesh and Gabriel Goh and Sandhini Agarwal and Girish Sastry and Amanda Askell and Pamela Mishkin and Jack Clark and Gretchen Krueger and Ilya Sutskever},
      year={2021},
      eprint={2103.00020},
      archivePrefix={arXiv},
      primaryClass={cs.CV},
      url={https://arxiv.org/abs/2103.00020}, 
}

@misc{girdhar2023imagebindembeddingspacebind,
      title={ImageBind: One Embedding Space To Bind Them All}, 
      author={Rohit Girdhar and Alaaeldin El-Nouby and Zhuang Liu and Mannat Singh and Kalyan Vasudev Alwala and Armand Joulin and Ishan Misra},
      year={2023},
      eprint={2305.05665},
      archivePrefix={arXiv},
      primaryClass={cs.CV},
      url={https://arxiv.org/abs/2305.05665}, 
}

@misc{zhu2024languagebindextendingvideolanguagepretraining,
      title={LanguageBind: Extending Video-Language Pretraining to N-modality by Language-based Semantic Alignment}, 
      author={Bin Zhu and Bin Lin and Munan Ning and Yang Yan and Jiaxi Cui and HongFa Wang and Yatian Pang and Wenhao Jiang and Junwu Zhang and Zongwei Li and Wancai Zhang and Zhifeng Li and Wei Liu and Li Yuan},
      year={2024},
      eprint={2310.01852},
      archivePrefix={arXiv},
      primaryClass={cs.CV},
      url={https://arxiv.org/abs/2310.01852}, 
}

@misc{elazar2021amnesicprobingbehavioralexplanation,
      title={Amnesic Probing: Behavioral Explanation with Amnesic Counterfactuals}, 
      author={Yanai Elazar and Shauli Ravfogel and Alon Jacovi and Yoav Goldberg},
      year={2021},
      eprint={2006.00995},
      archivePrefix={arXiv},
      primaryClass={cs.CL},
      url={https://arxiv.org/abs/2006.00995}, 
}

@inproceedings{meng2022locating,
  title={Locating and editing factual associations in gpt},
  author={Meng, Kevin and Bau, David and Andonian, Alex J and Belinkov, Yonatan},
  booktitle={Advances in neural information processing systems},
  year={2022}
}

@misc{wang2022interpretabilitywildcircuitindirect,
      title={Interpretability in the Wild: a Circuit for Indirect Object Identification in GPT-2 small}, 
      author={Kevin Wang and Alexandre Variengien and Arthur Conmy and Buck Shlegeris and Jacob Steinhardt},
      year={2022},
      eprint={2211.00593},
      archivePrefix={arXiv},
      primaryClass={cs.LG},
      url={https://arxiv.org/abs/2211.00593}, 
}

@article{conmy2023towards,
  title={Towards automated circuit discovery for mechanistic interpretability},
  author={Conmy, Arthur and Mavor-Parker, Augustine and Lynch, Aengus and Heimersheim, Stefan and Garriga-Alonso, Adri{\`a}},
  journal={Advances in Neural Information Processing Systems},
  volume={36},
  pages={16318--16352},
  year={2023}
}

@article{zhang2023towards,
  title={Towards best practices of activation patching in language models: Metrics and methods},
  author={Zhang, Fred and Nanda, Neel},
  journal={arXiv preprint arXiv:2309.16042},
  year={2023}
}

@inproceedings{ravfogel2020inlp,
  title = {Null It Out: Guarding Protected Attributes by Iterative Nullspace Projection},
  author = {Ravfogel, Shauli and Elazar, Yanai and Gonen, Hila and Twiton, Michael and Goldberg, Yoav},
  booktitle = {Proceedings of the 58th Annual Meeting of the Association for Computational Linguistics},
  pages = {7237--7256},
  year = {2020},
  doi = {10.18653/v1/2020.acl-main.647}
}

@article{belrose2023eliciting,
  title={Eliciting latent predictions from transformers with the tuned lens},
  author={Belrose, Nora and Ostrovsky, Igor and McKinney, Lev and Furman, Zach and Smith, Logan and Halawi, Danny and Biderman, Stella and Steinhardt, Jacob},
  journal={arXiv preprint arXiv:2303.08112},
  year={2023}
}

@inproceedings{pal2023future,
  title={Future lens: Anticipating subsequent tokens from a single hidden state},
  author={Pal, Koyena and Sun, Jiuding and Yuan, Andrew and Wallace, Byron C and Bau, David},
  booktitle={Proceedings of the 27th Conference on Computational Natural Language Learning (CoNLL)},
  pages={548--560},
  year={2023}
}

@misc{burns2024discoveringlatentknowledgelanguage,
      title={Discovering Latent Knowledge in Language Models Without Supervision}, 
      author={Collin Burns and Haotian Ye and Dan Klein and Jacob Steinhardt},
      year={2024},
      eprint={2212.03827},
      archivePrefix={arXiv},
      primaryClass={cs.CL},
      url={https://arxiv.org/abs/2212.03827}, 
}

@article{kadavath2022language,
  title={Language models (mostly) know what they know},
  author={Kadavath, Saurav and Conerly, Tom and Askell, Amanda and Henighan, Tom and Drain, Dawn and Perez, Ethan and Schiefer, Nicholas and Hatfield-Dodds, Zac and DasSarma, Nova and Tran-Johnson, Eli and others},
  journal={arXiv preprint arXiv:2207.05221},
  year={2022}
}

@article{marks2023geometry,
  title={The geometry of truth: Emergent linear structure in large language model representations of true/false datasets},
  author={Marks, Samuel and Tegmark, Max},
  journal={arXiv preprint arXiv:2310.06824},
  year={2023}
}

@article{burger2024truth,
  title={Truth is universal: Robust detection of lies in llms},
  author={B{\"u}rger, Lennart and Hamprecht, Fred A and Nadler, Boaz},
  journal={Advances in Neural Information Processing Systems},
  volume={37},
  pages={138393--138431},
  year={2024}
}

@inproceedings{azaria2023internal,
  title={The internal state of an LLM knows when it’s lying},
  author={Azaria, Amos and Mitchell, Tom},
  booktitle={Findings of the Association for Computational Linguistics: EMNLP 2023},
  pages={967--976},
  year={2023}
}

@article{alayrac2022flamingo,
  title={Flamingo: a visual language model for few-shot learning},
  author={Alayrac, Jean-Baptiste and Donahue, Jeff and Luc, Pauline and Miech, Antoine and Barr, Iain and Hasson, Yana and Lenc, Karel and Mensch, Arthur and Millican, Katie and Reynolds, Malcolm and others},
  journal={arXiv preprint arXiv:2204.14198},
  year={2022}
}

@article{liu2023visual,
  title={Visual instruction tuning},
  author={Liu, Haotian and Li, Chunyuan and Wu, Qingyang and Lee, Yong Jae},
  journal={Advances in neural information processing systems},
  volume={36},
  pages={34892--34916},
  year={2023}
}

@article{chu2024qwen2,
  title={Qwen2-audio technical report},
  author={Chu, Yunfei and Xu, Jin and Yang, Qian and Wei, Haojie and Wei, Xipin and Guo, Zhifang and Leng, Yichong and Lv, Yuanjun and He, Jinzheng and Lin, Junyang and others},
  journal={arXiv preprint arXiv:2407.10759},
  year={2024}
}

@inproceedings{zhan2024anygpt,
  title={Anygpt: Unified multimodal llm with discrete sequence modeling},
  author={Zhan, Jun and Dai, Junqi and Ye, Jiasheng and Zhou, Yunhua and Zhang, Dong and Liu, Zhigeng and Zhang, Xin and Yuan, Ruibin and Zhang, Ge and Li, Linyang and others},
  booktitle={Proceedings of the 62nd Annual Meeting of the Association for Computational Linguistics (Volume 1: Long Papers)},
  pages={9637--9662},
  year={2024}
}

@article{xu2025qwen3,
  title={Qwen3-omni technical report},
  author={Xu, Jin and Guo, Zhifang and Hu, Hangrui and Chu, Yunfei and Wang, Xiong and He, Jinzheng and Wang, Yuxuan and Shi, Xian and He, Ting and Zhu, Xinfa and others},
  journal={arXiv preprint arXiv:2509.17765},
  year={2025}
}

@article{team2026gemma,
  title={Gemma 4 technical report},
  author={Team, Gemma and Abd, Sherif El and Aggarwal, Vaibhav and Algayres, Robin and Andreev, Alek and Bachem, Olivier and Ballantyne, Ian and Brick, Cormac and C{\u{a}}rbune, Victor and Casbon, Michelle and others},
  journal={arXiv preprint arXiv:2607.02770},
  year={2026}
}

@article{cui2026minicpm,
  title={Minicpm-o 4.5: Towards real-time full-duplex omni-modal interaction},
  author={Cui, Junbo and Xu, Bokai and Wang, Chongyi and Yu, Tianyu and Sun, Weiyue and Xu, Yingjing and Wang, Tianran and He, Zhihui and Ma, Wenshuo and Cai, Tianchi and others},
  journal={arXiv preprint arXiv:2604.27393},
  year={2026}
}

@inproceedings{wang2026xmodbench,
  title={Xmodbench: Benchmarking cross-modal capabilities and consistency in omni-language models},
  author={Wang, Xingrui and Liu, Jiang and Huang, Chao and Yu, Xiaodong and Wang, Ze and Sun, Ximeng and Wu, Jialian and Yuille, Alan and Barsoum, Emad and Liu, Zicheng},
  booktitle={International Conference on Learning Representations},
  volume={2026},
  pages={19508--19521},
  year={2026}
}

\clearpage

\appendix

\section{Implementation Details}
\subsection{Representation Extraction}\label{app:extract}
For each item we render two prompts that differ only in the modality of the evidence and in the modality-specific wording of the question; the question stem, the four answer choices, and their assignment to the letters A--D are identical across the two conditions. 
The gold label never appears in the prompt.
Each prompt is passed through the model once with \texttt{output\_hidden\_states} enabled and no sampling. 
We record the hidden state at the position whose logits determine the answer letter, which is the final position of the rendered prompt, at every layer of the language backbone. 
The per-condition tensors are stored as \texttt{answer\_a} and \texttt{answer\_v} of shape $N\times L\times d$ together with the model's own predicted letter for each condition.
Prediction is read as the arg-max over the logits of the four letter tokens at that position, using the token id of each letter as it is tokenized in the answer slot. 
Items on which the answer letter cannot be resolved are dropped before any analysis; the retained counts appear in Table~\ref{tab:percond}.

\subsection{Answer Subspace Estimation}\label{app:readout}
At each layer we pool the answer-position states of both modality conditions and fit a multinomial logistic regression on standardized features (\texttt{StandardScaler} followed by \texttt{LogisticRegression}, \texttt{max\_iter}$=20000$, \texttt{tol}$=10^{-6}$, default $\ell_2$ penalty, fixed seed) with the model's own predicted letter as the target.
Cross-validation is grouped by item, so the two conditions of an item always fall in the same fold, and we report the mean over three schemes: stratified grouped $5$-fold, plain grouped $5$-fold, and stratified $5$-fold. All three give the same half-decodability depth in every condition; the spread across schemes is shown as the band in Figure~\ref{fig:fig1a}.
The coefficient matrix $W^{(l)}\in\mathds R^{d\times4}$ has effective rank at most three because softmax predictions are invariant to a constant shift of all class logits. 
We take a thin QR decomposition of $W^{(l)}$ and keep the first three columns of $Q$ as the orthonormal basis $Q_{\mathrm{ans}}^{(l)}$. 
Coordinates are obtained as $\mathbf h^\top Q_{\mathrm{ans}}^{(l)}$ and the complement by the projector in Eq.~\ref{eq:residual}.
Bases are re-estimated inside each fold; all reported accuracies are held-out.

\subsection{Disagreement Subspace Estimation}\label{app:dis}
The disagreement probe uses the same pipeline --- standardization, logistic regression with the same settings, item-grouped stratified $5$-fold --- with the binary label $z_i$ of Eq.~\ref{eq:disagreement} and one sample per item.
Its input is the concatenation of Eq.~\ref{eq:disfeat}, formed from the complement components $\mathbf r^{(m,l)}$ rather than the raw states, so no direction already used to decode the answer can enter. 
Reshaping the fitted coefficient vector into $W_z^{(l)}\in\mathds R^{3\times d}$ and taking the leading two right singular vectors gives $B_{\mathrm{dis}}^{(l)}$; the basis is re-orthogonalized against $Q_{\mathrm{ans}}^{(l)}$ and re-QR'd, which leaves a residual leakage into the Answer Subspace below $10^{-7}$ in every condition.
Alongside the two feature sets reported in the body we evaluate a rank-matched random readout of the complement, averaged over five draws, and a PCA readout of the same rank; these appear in Table~\ref{tab:loc}.

\subsection{Feature Blocks}\label{app:blocks}
The difference block $\mathbf r^{(\mathrm v)}-\mathbf r^{(\mathrm a)}$ is a linear function of the other two and therefore does not enlarge the hypothesis class of a linear probe. 
We keep it for a different reason: with three blocks the reshaped coefficient matrix has rank at most three, so saturation at two directions is an observation about the data. 
With two blocks the rank ceiling would be exactly two and the same fact would be a construction artifact.
Table~\ref{tab:blocks} confirms that the choice does not change the result. 
Discriminability with two blocks matches the three-block value to within $0.019$, and the planes recovered under the two parameterizations overlap at $0.972$--$0.991$.

\begin{table}[t]
    \centering
    \small
    \setlength{\tabcolsep}{4pt}
    \begin{tabular}{llrrr}
        \hline
        Model & Family & 3 blocks & 2 blocks & Overlap \\
        \hline
        Qwen2.5-Omni & Ext. & 0.679 & 0.674 & 0.972 \\
        Qwen2.5-Omni & Spa. & 0.704 & 0.687 & 0.985 \\
        Gemma-4-E4B & Ext. & 0.718 & 0.712 & 0.988 \\
        Gemma-4-E4B & Spa. & 0.681 & 0.662 & 0.989 \\
        MiniCPM-o 4.5 & Ext. & 0.723 & 0.723 & 0.978 \\
        MiniCPM-o 4.5 & Spa. & 0.639 & 0.638 & 0.991 \\
        \hline
    \end{tabular}
    \caption{
        Effect of removing the difference block. The first two columns are balanced accuracy of the disagreement probe on the full complement; \emph{Overlap} is the mean principal-angle cosine between the two-dimensional subspaces recovered under the two parameterizations.
    }
    \label{tab:blocks}
\end{table}

\subsection{Intervention Details}\label{app:interv}
All interventions add a fixed vector to the answer-position state at a single layer during a forward pass, as in Eq.~\ref{eq:interv}, and leave every other position and layer untouched.

\paragraph{Direction.}
For the Answer Subspace, $\boldsymbol\delta$ points from the centroid of the currently decoded answer region to the centroid of the next letter in cyclic order, so the perturbation targets a specific alternative rather than an arbitrary direction. For the Disagreement-Predictive Subspace, $\boldsymbol\delta$ is the difference between the centroids of disagreement and agreement items in the two-dimensional coordinates. For the random control, $\boldsymbol\delta$ is an isotropic Gaussian draw with a fixed seed.

\paragraph{Normalization.}
Each $\boldsymbol\delta$ is rescaled to unit length and multiplied by the coordinate standard deviation of its own subspace, so one unit of $\alpha$ is one standard deviation of that subspace's coordinates. This matters because the two subspaces differ in both dimensionality and scale: at the intervention depth of Qwen2.5-Omni on \emph{external}, the coordinate spread is $7.24$ in the Answer Subspace and $0.84$ in the Disagreement-Predictive Subspace, so an unnormalized comparison would favor one arm by roughly a factor of nine.

\paragraph{Controls.}
The random control is a two-dimensional subspace drawn from the orthogonal complement of the Answer Subspace and dimension-matched to the Disagreement-Predictive Subspace. The pre-formation control applies the identical operation at a shallower layer, chosen per condition to fall before the depth at which the answer first becomes decodable.

\paragraph{Depths.}
Interventions use a layer at approximately $75\%$ of normalized depth; the exact layer and the pre-formation layer for each condition appear in Table~\ref{tab:percond}.

\section{Per-condition Results}
\subsection{Conditions, Depths and Cohort Sizes}\label{app:percond}
Table~\ref{tab:percond} lists every condition used in the paper, together with the depths at which the answer first becomes decodable and at which interventions are applied.

\begin{table*}[t]
    \centering
    \small
    \begin{tabular}{llrrrrrrrr}
        \hline
        Model & Family & Layers & Items & Used & Agree & Disagree & Half-dec.\ layer & Half-dec.\ depth & Interv.\ depth \\
        \hline
        Qwen2.5-Omni & Ext. & 29 & 1350 & 1350 & 1038 & 312 & 15 & 53.600 & 75.000 \\
        Qwen2.5-Omni & Spa. & 29 & 465 & 465 & 165 & 300 & 19 & 67.900 & 75.000 \\
        Gemma-4-E4B & Ext. & 43 & 1350 & 1350 & 528 & 822 & 18 & 42.900 & 76.200 \\
        Gemma-4-E4B & Spa. & 43 & 465 & 465 & 213 & 252 & 24 & 57.100 & 76.200 \\
        MiniCPM-o 4.5 & Ext. & 37 & 1350 & 1350 & 864 & 486 & 17 & 47.200 & 75.000 \\
        MiniCPM-o 4.5 & Spa. & 37 & 465 & 465 & 141 & 324 & 20 & 55.600 & 75.000 \\
        \hline
    \end{tabular}
    \caption{
        Conditions used throughout. \emph{Used} is the number of items on which the answer letter could be resolved in both conditions; \emph{Agree} and \emph{Disagree} split those items by whether the two modality conditions produced the same letter. \emph{Half-dec.\ layer} is the layer at which answer-probe accuracy first reaches the midpoint between chance and its final-layer value, and \emph{Half-dec.\ depth} the corresponding normalized depth. \emph{Interv.\ depth} is the normalized depth of the layer at which interventions are applied.
    }
    \label{tab:percond}
\end{table*}

\subsection{Localization of Disagreement-Predictive Information}\label{app:loc}
Table~\ref{tab:loc} gives the per-condition values behind Figure~\ref{fig:fig1b}, evaluated at the final layer; interventions instead use the depths in Table~\ref{tab:percond}.

\begin{table*}[t]
    \centering
    \small
    \setlength{\tabcolsep}{4pt}
    \begin{tabular}{llrrrrr}
        \hline
        Model & Family & Inside & Compl.\ & Random & PCA & Full \\
        \hline
        Qwen2.5-Omni & Ext. & 0.510 & 0.704 & 0.511 & 0.518 & 0.705 \\
        Qwen2.5-Omni & Spa. & 0.516 & 0.692 & 0.510 & 0.517 & 0.706 \\
        Gemma-4-E4B & Ext. & 0.586 & 0.692 & 0.534 & 0.623 & 0.699 \\
        Gemma-4-E4B & Spa. & 0.645 & 0.687 & 0.572 & 0.631 & 0.692 \\
        MiniCPM-o 4.5 & Ext. & 0.500 & 0.725 & 0.679 & 0.685 & 0.718 \\
        MiniCPM-o 4.5 & Spa. & 0.495 & 0.659 & 0.516 & 0.544 & 0.651 \\
        \hline
    \end{tabular}
    \caption{
        Balanced accuracy of the disagreement probe at the final layer under five feature sets: coordinates inside the Answer Subspace, inside its orthogonal complement, a rank-matched random readout of that complement (mean of five draws), a rank-matched PCA readout of the complement, and the full hidden state. Chance is $0.5$. The complement matches the full state to within $0.014$ in every condition, while the two rank-matched readouts of the same complement fall well short of it, so the result does not follow from the complement's dimensionality alone.
    }
    \label{tab:loc}
\end{table*}

\subsection{Dimensionality and Stability}\label{app:stab}
Directions are estimated separately within each training fold and evaluated on the corresponding held-out fold.
Because the reshaped coefficient matrix $W_z^{(l)}\in\mathds R^{3\times d}$ has rank at most three, we evaluate the leading one, two, and three right singular vectors.
Table~\ref{tab:stab} reports how disagreement discriminability changes with the number of retained directions and how stable the leading two-dimensional subspace is under item-level resampling.

\begin{table*}[t]
    \centering
    \small
    \begin{tabular}{llrrrrrrr}
        \hline
        Model & Family & $k{=}1$ & $k{=}2$ & $k{=}3$ & Random $k{=}2$ & Bootstrap align.\ & Min & Null \\
        \hline
        Qwen2.5-Omni & Ext. & 0.593 & 0.692 & 0.694 & 0.515 & 0.733 & 0.683 & 0.033 \\
        Qwen2.5-Omni & Spa. & 0.666 & 0.694 & 0.691 & 0.512 & 0.725 & 0.632 & 0.031 \\
        Gemma-4-E4B & Ext. & 0.649 & 0.670 & 0.664 & 0.535 & 0.713 & 0.674 & 0.037 \\
        Gemma-4-E4B & Spa. & 0.629 & 0.660 & 0.674 & 0.566 & 0.719 & 0.670 & 0.039 \\
        MiniCPM-o 4.5 & Ext. & 0.625 & 0.721 & 0.718 & 0.568 & 0.724 & 0.693 & 0.029 \\
        MiniCPM-o 4.5 & Spa. & 0.611 & 0.639 & 0.641 & 0.507 & 0.747 & 0.700 & 0.028 \\
        \hline
    \end{tabular}
    \caption{
        Number of retained directions and stability of the recovered plane. The first four columns are balanced accuracy of the disagreement probe on $k$ directions of the complement, and on a random two-dimensional subspace of the same complement. \emph{Bootstrap align.}\ is the mean principal-angle cosine between the plane recovered from an item-level bootstrap resample ($30$ resamples) and the plane recovered from the full sample; \emph{Min} is the smallest such value and \emph{Null} the $95$th percentile obtained when the resampled directions are replaced by random ones.
    }
    \label{tab:stab}
\end{table*}

\subsection{Answer Recovery under the Original Prompt}\label{app:recover}
Table~\ref{tab:recover} gives the values behind the recovery accuracy quoted in Section~\ref{sec:results}.

\begin{table*}[t]
    \centering
    \small
    \setlength{\tabcolsep}{4pt}
    \begin{tabular}{llrrrr}
        \hline
         & & \multicolumn{2}{c}{Agree} & \multicolumn{2}{c}{Disagree} \\
        \cline{3-4}\cline{5-6}
        Model & Family & Aud.\ & Img.\ & Aud.\ & Img.\ \\
        \hline
        Qwen2.5-Omni & Ext. & 0.992 & 0.990 & 0.978 & 0.955 \\
        Qwen2.5-Omni & Spa. & 0.939 & 0.982 & 0.970 & 0.997 \\
        Gemma-4-E4B & Ext. & 0.987 & 0.991 & 0.987 & 0.996 \\
        Gemma-4-E4B & Spa. & 0.967 & 0.981 & 0.937 & 0.952 \\
        MiniCPM-o 4.5 & Ext. & 0.983 & 0.987 & 0.949 & 0.969 \\
        MiniCPM-o 4.5 & Spa. & 0.979 & 0.993 & 0.954 & 0.985 \\
        \hline
    \end{tabular}
    \caption{
        Agreement between the answer decoded from the final-layer Answer Subspace and the model's actual prediction, reported separately for each modality condition and for items where the two conditions agree or disagree. Agreement exceeds $0.93$ throughout, including items where the two conditions produced different letters.
    }
    \label{tab:recover}
\end{table*}

\subsection{Option Permutation}\label{app:permtab}
Table~\ref{tab:permtab} decomposes the permutation experiment by what the model did, showing that the decoded answer follows the model in every behavior group.

\begin{table*}[t]
    \centering
    \small
    \begin{tabular}{llrrrrrrr}
        \hline
         & & & \multicolumn{3}{c}{Model behavior} & \multicolumn{3}{c}{Decoded answer agrees with} \\
        \cline{4-6}\cline{7-9}
        Model & Family & $N$ & Content & Letter & Other & Model & Model\,$\mid$\,content & Model\,$\mid$\,letter \\
        \hline
        Qwen2.5-Omni & Ext. & 2700 & 0.919 & 0.028 & 0.053 & 0.985 & 0.991 & 0.895 \\
        Qwen2.5-Omni & Spa. & 1320 & 0.618 & 0.175 & 0.207 & 0.948 & 0.971 & 0.935 \\
        Gemma-4-E4B & Ext. & 2700 & 0.632 & 0.184 & 0.184 & 0.984 & 0.989 & 0.978 \\
        Gemma-4-E4B & Spa. & 1320 & 0.366 & 0.305 & 0.329 & 0.967 & 0.969 & 0.970 \\
        MiniCPM-o 4.5 & Ext. & 2700 & 0.880 & 0.043 & 0.076 & 0.985 & 0.990 & 0.966 \\
        MiniCPM-o 4.5 & Spa. & 1320 & 0.777 & 0.081 & 0.142 & 0.989 & 0.997 & 0.963 \\
        \hline
    \end{tabular}
    \caption{
        Option permutation, decomposed. $N$ is the number of (item, modality) pairs. \emph{Model behavior} is how the model's own prediction moved: to the letter now holding the original content, staying on the original letter, or to a third letter. The remaining columns are the fraction of pairs on which the answer decoded from the Answer Subspace agrees with the model's new prediction, overall and restricted to the two behavior groups. The readout follows the model in every group, including when the model stays on the original letter.
    }
    \label{tab:permtab}
\end{table*}

\subsection{Geometry Relative to Modality}\label{app:geotab}
Table~\ref{tab:geotab} gives the raw pair distances summarized in Section~\ref{sec:results} and the per-condition angles behind Figure~\ref{fig:modctrl}.

\begin{table*}[t]
    \centering
    \small
    \begin{tabular}{llrrrrrrr}
        \hline
         & & \multicolumn{3}{c}{Pair distance} & & \multicolumn{3}{c}{Modality} \\
        \cline{3-5}\cline{7-9}
        Model & Family & Agree & Disagree & Ratio & Excess/dim & Separability & $\cos$ answer & $\cos$ disagree \\
        \hline
        Qwen2.5-Omni & Ext. & 0.170 & 0.225 & 1.324 & 7.310 & 1.000 & 0.054 & 0.023 \\
        Qwen2.5-Omni & Spa. & 0.173 & 0.218 & 1.260 & 3.170 & 1.000 & 0.034 & 0.023 \\
        Gemma-4-E4B & Ext. & 0.681 & 0.931 & 1.367 & 8.620 & 1.000 & 0.065 & 0.036 \\
        Gemma-4-E4B & Spa. & 0.841 & 0.899 & 1.069 & 4.750 & 1.000 & 0.040 & 0.060 \\
        MiniCPM-o 4.5 & Ext. & 0.892 & 1.060 & 1.188 & 2.920 & 1.000 & 0.046 & 0.017 \\
        MiniCPM-o 4.5 & Spa. & 1.266 & 1.316 & 1.039 & 3.070 & 1.000 & 0.010 & 0.043 \\
        \hline
    \end{tabular}
    \caption{
        Geometry at the intervention depth. \emph{Pair distance} is the mean distance between the two modality conditions of the same item inside the Answer Subspace, split by cohort, with their ratio. \emph{Excess/dim} is how much more the excess separation of disagreeing pairs concentrates inside the Answer Subspace than in its complement, per dimension. \emph{Separability} is the balanced accuracy of a probe predicting which modality the input arrived in from the answer-position state; the two $\cos$ columns are the cosine of the principal angle between the single axis that probe defines and each subspace. Removing the mean modality offset changes disagreement accuracy by at most $0.0031$.
    }
    \label{tab:geotab}
\end{table*}

\subsection{Answer-matched Control}\label{app:ansmatch}
Table~\ref{tab:ansmatch} reports the strata, sample sizes and unrestricted values for the control shown in Figure~\ref{fig:ansctrl}.

\begin{table*}[t]
    \centering
    \small
    \setlength{\tabcolsep}{4pt}
    \begin{tabular}{llrrrrr}
        \hline
        Model & Family & Strata & $N$ & Unrestr.\ & Matched & Null \\
        \hline
        Qwen2.5-Omni & Ext. & 4 & 675 & 0.635 & 0.670 & 0.509 \\
        Qwen2.5-Omni & Spa. & 4 & 233 & 0.558 & 0.726 & 0.587 \\
        Gemma-4-E4B & Ext. & 4 & 675 & 0.642 & 0.712 & 0.528 \\
        Gemma-4-E4B & Spa. & 2 & 157 & 0.648 & 0.703 & 0.586 \\
        MiniCPM-o 4.5 & Ext. & 4 & 675 & 0.683 & 0.697 & 0.534 \\
        MiniCPM-o 4.5 & Spa. & 2 & 171 & 0.605 & 0.711 & 0.524 \\
        \hline
    \end{tabular}
    \caption{
        Holding the audio-condition prediction fixed. A fully answer-pair-matched comparison is impossible because a fixed pair of predictions determines the disagreement label, so we stratify on the audio-condition prediction alone and pool the within-stratum results. \emph{Strata} is the number of audio-condition answers that contain both cohorts, $N$ the number of items retained, and \emph{Null} the $95$th percentile of a label-permutation null computed under the same stratification. The matched values exceed that null in every condition and are no lower than the unrestricted ones; the unrestricted value for Qwen2.5-Omni on \emph{spatial} falls below the null, which is why the body claim rests on the matched series.
    }
    \label{tab:ansmatch}
\end{table*}

\subsection{Pair Displacement}\label{app:pairs}
Table~\ref{tab:pairdist} gives the pair distances computed in the full subspace rather than in the projection shown in Figure~\ref{fig:fig7}.

\begin{table*}[t]
    \centering
    \small
    \setlength{\tabcolsep}{4pt}
    \begin{tabular}{llrrrrr}
        \hline
        Model & Family & $N$ & Disagree & Dist.\ disagree & Dist.\ agree & Ratio \\
        \hline
        Qwen2.5-Omni & Ext. & 1350 & 312 & 2.199 & 0.418 & 5.263 \\
        Qwen2.5-Omni & Spa. & 465 & 300 & 2.694 & 1.117 & 2.413 \\
        Gemma-4-E4B & Ext. & 1350 & 822 & 2.877 & 0.573 & 5.021 \\
        Gemma-4-E4B & Spa. & 465 & 252 & 2.247 & 0.913 & 2.460 \\
        MiniCPM-o 4.5 & Ext. & 1350 & 486 & 2.113 & 0.896 & 2.358 \\
        MiniCPM-o 4.5 & Spa. & 465 & 324 & 2.615 & 1.362 & 1.921 \\
        \hline
    \end{tabular}
    \caption{
        Median distance between the two modality conditions of the same item, computed in the full three-dimensional Answer Subspace rather than in the two-dimensional projection shown in Figure~\ref{fig:fig7}, in units of the coordinate standard deviation. \emph{Disagree} is the number of disagreement items among $N$; \emph{Dist.\ disagree} and \emph{Dist.\ agree} are the median pair distances within each cohort and \emph{Ratio} their quotient.
    }
    \label{tab:pairdist}
\end{table*}

\section{Intervention Results}
\subsection{Magnitude Sweep}\label{app:dose}
Table~\ref{tab:dose} gives the per-condition values behind Figure~\ref{fig:fig8}.

\begin{table*}[t]
    \centering
    \small
    \setlength{\tabcolsep}{5pt}
    \begin{tabular}{llrrrrrr}
        \hline
        Subspace & $\alpha$
        & Qwen Ext. & Qwen Spa.
        & Gemma Ext. & Gemma Spa.
        & MiniCPM Ext. & MiniCPM Spa. \\
        \hline
        \multirow{5}{*}{Answer}
        & 2  & 0.040 & 0.065 & 0.020 & 0.110 & 0.105 & 0.085 \\
        & 5  & 0.090 & 0.135 & 0.075 & 0.275 & 0.180 & 0.195 \\
        & 10 & 0.170 & 0.250 & 0.295 & 0.500 & 0.740 & 0.810 \\
        & 20 & 0.640 & 0.485 & 0.860 & 0.760 & 0.900 & 0.995 \\
        & 40 & 0.980 & 0.895 & 0.980 & 0.825 & 1.000 & 0.995 \\
        \hline
        \multirow{5}{*}{Disagreement}
        & 2  & 0.000 & 0.005 & 0.000 & 0.010 & 0.005 & 0.005 \\
        & 5  & 0.005 & 0.000 & 0.000 & 0.010 & 0.005 & 0.010 \\
        & 10 & 0.005 & 0.005 & 0.000 & 0.020 & 0.000 & 0.005 \\
        & 20 & 0.000 & 0.015 & 0.000 & 0.030 & 0.005 & 0.015 \\
        & 40 & 0.015 & 0.025 & 0.000 & 0.050 & 0.015 & 0.030 \\
        \hline
        \multirow{5}{*}{Random}
        & 2  & 0.005 & 0.005 & 0.000 & 0.005 & 0.005 & 0.005 \\
        & 5  & 0.000 & 0.000 & 0.000 & 0.010 & 0.005 & 0.005 \\
        & 10 & 0.000 & 0.000 & 0.000 & 0.015 & 0.005 & 0.010 \\
        & 20 & 0.005 & 0.005 & 0.005 & 0.045 & 0.005 & 0.010 \\
        & 40 & 0.005 & 0.010 & 0.010 & 0.090 & 0.010 & 0.015 \\
        \hline
    \end{tabular}
    \caption{
        Fraction of predictions changed across intervention magnitudes $\alpha$,
        measured in target-subspace standard deviations, over $100$ items per
        condition. Answer Subspace interventions increase monotonically and
        exceed both controls from $\alpha=2$ onward.
    }
    \label{tab:dose}
\end{table*}

\subsection{Prediction Changes by Subspace and Depth}\label{app:intervtab}
Table~\ref{tab:intervtab} gives the per-condition counts, including the pre-formation control.

\begin{table*}[t]
    \centering
    \small
    \begin{tabular}{llrrrrcrr}
        \hline
         & & \multicolumn{4}{c}{At intervention depth} & & \multicolumn{2}{c}{Before answer forms} \\
        \cline{3-6}\cline{8-9}
         & & \multicolumn{2}{c}{Answer} & Disagree & Random & & Answer & Disagree \\
        Model & Family & $\alpha{=}10$ & $\alpha{=}40$ & $\alpha{=}40$ & $\alpha{=}40$ & & $\alpha{=}40$ & $\alpha{=}40$ \\
        \hline
        Qwen2.5-Omni & Ext. & 28 & 98 & 1 & 0 & & 0 & 0 \\
        Qwen2.5-Omni & Spa. & 45 & 96 & 6 & 2 & & 1 & 2 \\
        Gemma-4-E4B & Ext. & 36 & 100 & 0 & 1 & & 1 & 0 \\
        Gemma-4-E4B & Spa. & 56 & 89 & 6 & 18 & & 2 & 9 \\
        MiniCPM-o 4.5 & Ext. & 75 & 100 & 1 & 3 & & 0 & 1 \\
        MiniCPM-o 4.5 & Spa. & 85 & 99 & 5 & 4 & & 3 & 4 \\
        \hline
    \end{tabular}
    \caption{
        Number of predictions changed out of $100$ items. All arms use the same layer, operation and magnitude within a depth group. Perturbing the Answer Subspace at the intervention depth changes $89$--$100$ predictions, and the same perturbation applied before the answer becomes decodable changes $0$--$3$. The Disagreement-Predictive Subspace and the random control are both near zero and show no consistent ordering: the random control exceeds the disagreement subspace in three of the six conditions, most visibly for Gemma-4-E4B on \emph{spatial}, where a random two-dimensional subspace changes $18$ predictions. In that condition the two arms are therefore not distinguishable and the claim rests on the gap against the Answer Subspace rather than on an ordering between the controls.
    }
    \label{tab:intervtab}
\end{table*}

\subsection{Cohort Split}\label{app:cohort}
Table~\ref{tab:cohort} splits the Answer Subspace intervention by whether the two modality conditions originally agreed.

\begin{table*}[t]
    \centering
    \small
    \begin{tabular}{llrrrrrcrrr}
        \hline
         & & \multicolumn{2}{c}{$N$} & \multicolumn{3}{c}{Changed at $\alpha{=}10$} & & \multicolumn{3}{c}{Settling depth (\%)} \\
        \cline{3-4}\cline{5-7}\cline{9-11}
        Model & Family & Agree & Disagree & Agree & Disagree & $\Delta$ & & Agree & Disagree & $\Delta$ \\
        \hline
        Qwen2.5-Omni & Ext. & 69 & 31 & 0.159 & 0.548 & 0.389 & & 49.900 & 57.300 & 7.400 \\
        Qwen2.5-Omni & Spa. & 34 & 66 & 0.471 & 0.439 & -0.031 & & 59.600 & 59.600 & 0.000 \\
        Gemma-4-E4B & Ext. & 59 & 41 & 0.390 & 0.317 & -0.073 & & 50.800 & 49.900 & -0.900 \\
        Gemma-4-E4B & Spa. & 47 & 53 & 0.426 & 0.679 & 0.254 & & 58.200 & 62.900 & 4.700 \\
        MiniCPM-o 4.5 & Ext. & 70 & 30 & 0.714 & 0.833 & 0.119 & & 48.000 & 52.900 & 4.900 \\
        MiniCPM-o 4.5 & Spa. & 36 & 64 & 0.778 & 0.891 & 0.113 & & 53.600 & 55.300 & 1.700 \\
        \hline
    \end{tabular}
    \caption{
        Answer Subspace intervention split by cohort, at the smaller magnitude. Disagreement items change more often in four of the six conditions, and the sign of that difference agrees with the difference in settling depth --- the normalized depth from which the decoded answer stops changing --- in every condition ($\rho=0.94$, $p=0.005$). At $\alpha=40$ both cohorts are near ceiling and the difference disappears. Cohort sizes are unequal because the split follows the model's own behavior, so these comparisons are the least powered in the paper and we do not build any claim in the body on them.
    }
    \label{tab:cohort}
\end{table*}

\subsection{Removal Interventions}\label{app:removal}
Instead of shifting the state along a direction, a removal intervention projects the state onto the orthogonal complement of the target subspace, deleting that subspace's contribution altogether. We report it separately because its outcome is weaker and more ambiguous than the shift intervention, and because the ambiguity is informative.
Table~\ref{tab:removal} gives the change in the margin between the model's top two answer letters after removing (i) the two-dimensional disagreement subspace, (ii) the three-dimensional Answer Subspace, and (iii) random subspaces of each dimensionality, over $240$ (item, modality) pairs per cell with paired bootstrap intervals.
Removing the Answer Subspace moves the margin significantly in five of the six conditions, and most strongly for MiniCPM-o 4.5 on \emph{external} ($-0.416$). Removing the disagreement subspace is indistinguishable from zero in five conditions and marginally significant in one ($-0.014$); in that same condition a random two-dimensional subspace is also significant and of comparable magnitude ($-0.013$), so the effect is not attributable to the disagreement directions.
Two limits keep us from resting any claim on this experiment. First, the removed energy differs across arms: a two-dimensional removal deletes $0.5$--$5.3\%$ of the state norm while removing the three-dimensional Answer Subspace deletes $3.4$--$8.5\%$, so the arms are not matched in the way the shift intervention is. Second, the number of outright prediction flips is small in every arm ($0$--$9$ of $240$), so the margin is the only sensitive readout and does not directly measure discrete prediction changes. The shift intervention reported in the body avoids both problems.

\begin{table*}[t]
    \centering
    \small
    \begin{tabular}{llrrrrr}
        \hline
        Model & Family & Disagree ($k{=}2$) & Random ($k{=}2$) & Random ($k{=}2$, b) & Answer ($k{=}3$) & Random ($k{=}3$) \\
        \hline
        Qwen2.5-Omni & Ext. & -0.002 & +0.002 & +0.008 & \textbf{-0.103} & \textbf{-0.033} \\
        Qwen2.5-Omni & Spa. & +0.002 & -0.000 & -0.000 & \textbf{-0.041} & -0.009 \\
        Gemma-4-E4B & Ext. & +0.008 & \textbf{+0.043} & \textbf{-0.027} & \textbf{-0.054} & +0.006 \\
        Gemma-4-E4B & Spa. & +0.003 & -0.004 & -0.007 & \textbf{+0.039} & -0.002 \\
        MiniCPM-o 4.5 & Ext. & \textbf{-0.015} & +0.006 & \textbf{-0.013} & \textbf{-0.416} & \textbf{-0.019} \\
        MiniCPM-o 4.5 & Spa. & -0.005 & -0.001 & \textbf{-0.011} & -0.027 & \textbf{-0.018} \\
        \hline
    \end{tabular}
    \caption{
        Change in the top-two answer margin after removing each subspace, over $240$ (item, modality) pairs. Values in bold differ from zero at the $95\%$ level under a paired bootstrap. Two independent random draws at $k=2$ are shown to indicate the spread of the control itself.
    }
    \label{tab:removal}
\end{table*}

\subsection{Verification that Interventions Reach Their Target}\label{app:injdetail}
A null behavioral result is only interpretable if the intervention actually moved the state where it was aimed. We verified this on Qwen2.5-Omni, \emph{external}, over $40$ items, by reading the post-intervention state back out at the intervention layer.
Injecting $\alpha=40$ produced a displacement of $39.97$ coordinate standard deviations along the target direction, so the intended shift is delivered essentially exactly. The disagreement probe called the perturbed state a disagreement item in every case, confirming that the state was moved into the region the probe associates with disagreement. No prediction changed.
Two observations qualify this. Measured at the deeper layers where the answer is read, the displacement has shrunk to $3.88$ standard deviations, and after the final layer normalization the residual displacement is $0.87$ standard deviations. We report these as observations only: whether the shrinkage reflects an active correction or the ordinary contraction of an off-distribution perturbation is not something this measurement distinguishes. The verification establishes that the intervention reached its target at the layer where it was applied, which is what the null result requires; it does not establish that the perturbation survives to the output unchanged.

\subsection{Asymmetric Intervention across Magnitudes}\label{app:asymsweep}
The asymmetric experiment in the body uses a single magnitude. Table~\ref{tab:asymsweep} reports an independent rerun of both arms over the full magnitude grid. The result bounds the claim: the asymmetric intervention exceeds its symmetric control in every condition at $\alpha=40$, in five of six at $\alpha=20$, and in neither direction consistently below that. At $\alpha\leq5$ the symmetric control is in fact higher in every condition.

The mechanism is straightforward. Creating disagreement asymmetrically requires the perturbed condition's own answer to move, and at $\alpha=2$ the Answer Subspace intervention changes only $0.071$ of predictions, so most items are untouched in both conditions. The symmetric arm has a different failure mode: applying the same shift to both conditions moves their predictions at different magnitudes, and that incidental desynchronization produces disagreement more often than a deliberate asymmetry does when neither condition is reliably displaced. The experiment therefore presupposes a magnitude at which the answer representation can actually be relocated.

One observation holds at every magnitude: the prediction of the untouched condition never moved, in any of the $1800$ item-by-magnitude combinations.

\begin{table*}[t]
    \centering
    \small
    \begin{tabular}{llrrrrrrrrrrr}
        \hline
         & & & \multicolumn{5}{c}{Asymmetric} & \multicolumn{5}{c}{Symmetric control} \\
        \cline{4-8}\cline{9-13}
        Model & Family & $N$ & $2$ & $5$ & $10$ & $20$ & $40$ & $2$ & $5$ & $10$ & $20$ & $40$ \\
        \hline
        Qwen2.5-Omni & Ext. & 60 & 0.000 & 0.000 & 0.017 & 0.500 & 0.983 & 0.033 & 0.083 & 0.183 & 0.300 & 0.050 \\
        Qwen2.5-Omni & Spa. & 60 & 0.033 & 0.033 & 0.100 & 0.367 & 0.833 & 0.200 & 0.300 & 0.550 & 0.600 & 0.233 \\
        Gemma-4-E4B & Ext. & 60 & 0.000 & 0.000 & 0.200 & 0.817 & 0.950 & 0.033 & 0.133 & 0.383 & 0.183 & 0.083 \\
        Gemma-4-E4B & Spa. & 60 & 0.000 & 0.217 & 0.400 & 0.700 & 0.767 & 0.067 & 0.283 & 0.433 & 0.550 & 0.250 \\
        MiniCPM-o 4.5 & Ext. & 60 & 0.017 & 0.033 & 0.717 & 0.950 & 1.000 & 0.083 & 0.217 & 0.233 & 0.200 & 0.100 \\
        MiniCPM-o 4.5 & Spa. & 60 & 0.050 & 0.100 & 0.867 & 1.000 & 1.000 & 0.133 & 0.333 & 0.333 & 0.067 & 0.067 \\
        \hline
    \end{tabular}
    \caption{
        Fraction of originally agreeing items on which disagreement appears, by intervention magnitude $\alpha$ in units of the coordinate standard deviation. \emph{Asymmetric} perturbs the Answer Subspace under one modality only; the symmetric control applies the identical shift to both. $N$ is the number of originally agreeing items evaluated. The asymmetric arm exceeds the control in all six conditions at $\alpha=40$ (by at least a factor of $3.07$), in five of six at $\alpha=20$, and in none at $\alpha\leq5$.
    }
    \label{tab:asymsweep}
\end{table*}

\end{document}